# Redundancy-as-Masking: Formalizing the Artificial Age Score (AAS) to Model Memory Aging in Generative AI

Seyma Yaman Kayadibi
Victoria University
seyma.yamankayadibi@live.vu.edu.au

## Abstract

Artificial intelligence can exhibit aging-like patterns not as a function of chronological time, but through systematic asymmetries in output-level observable memory performance under different context-persistence conditions. In large language models, semantic cues, such as the name of the day, often remain stable across sessions, while episodic details, like the sequential progression of experiment numbers, tend to collapse when conversational context is reset in the 25-day bilingual recall protocol. To capture this phenomenon, the Artificial Age Score (AAS) is introduced as a log-scaled, entropy-informed metric of memory age derived from observable recall behavior and defined purely at the output level, without access to internal latent-state representations. The score is formally proven to be well-defined, bounded, and monotonic under mild and model-agnostic assumptions, supporting mathematical reuse across evaluation settings while not implying empirical generalization across models or domains. In its Redundancy-as-Masking formulation, the score interprets redundancy as overlapping information that reduces the penalized mass. However, in the present study, redundancy is not explicitly estimated; all reported values assume a redundancy-neutral setting ($R = 0$), yielding conservative upper bounds.

The AAS framework was tested over a 25-day bilingual study involving ChatGPT-5.0, structured into stateless and persistent interaction phases. During persistent sessions, the model consistently recalled both semantic and episodic details, driving the AAS toward its theoretical minimum, indicative of behavioural youth in recall. In contrast, when sessions were reset, the model preserved semantic consistency but failed to maintain episodic continuity, causing a sharp increase in the AAS and signaling an aging-like behavioural signature of continuity loss in recall behavior. These results are interpreted behaviorally and do not constitute evidence about internal memory mechanisms or latent-state dynamics. These findings support the utility of AAS as a theoretically grounded, task-independent diagnostic tool for evaluating memory degradation in artificial systems. The empirical validation reported here is limited to this specific model version and protocol; applications to other architectures or training regimes require revalidation. The study builds on foundational concepts from von Neumann's work on automata, Shannon's theories of information and redundancy, and Turing's behavioral approach to intelligence.

## 1. Introduction

In large-scale computational systems, decline is not indexed by chronological time but by the weakening of memory organization, the accumulation of repetitive operations, and distortions in information flow. Here, "decline" and "aging" are used in an operational sense to denote systematic changes in observable recall behavior under controlled interaction regimes, rather than claims about internal representations or mechanisms. A systems-level view of limitation was articulated in reflections on the brain and the computer, where system behavior was linked to constraints on storage and signal processing rather than to temporal aging alone (von Neumann, 1958/2012). Within this perspective, competence is appropriately assessed through observable performance when internal states are inaccessible, a behavioral stance established by the imitation game, which redirected the evaluation of intelligence from hidden mechanisms to external responses (Turing, 1950). In parallel, a quantitative account of information was formulated from symbol statistics rather than semantic content, thereby enabling measurement without access to internal meaning (Shannon, 1948). This lineage motivates a metric approach to memory aging in which surface behavior is used as evidence when mechanisms are opaque. The Artificial Age Score (AAS) is introduced as a theorem-based, output-level behavioural metric of memory age, defined solely in terms of observable recall scores and redundancy values rather than internal latent states. By construction, AAS depends only on these observable quantities; any two systems that produce the same pattern of recall scores and redundancy values will therefore receive the same AAS, even if their internal configurations differ. The score employs a logarithmic penalty kernel that vanishes under perfect recall and grows smoothly as recall deteriorates. Under mild assumptions, three properties are established: the score is well-defined and decomposable, each term is finite and regrouping does not affect totals; it is globally bounded; and it is monotone, penalties decrease as recall improves, and increase with task weights. Under the Redundancy-as-Masking interpretation, redundancy is treated as information overlap that discounts penalized mass through a multiplicative factor, reflecting the information-theoretic observation that repeated outputs contain less novel content (Shannon, 1948, 1951). Shannon-type quantities summarize output-level symbol statistics; the link between entropy redundancy and “aging” is not mechanistic: in this work, entropy-informed penalization is used to operationalize an aging-like signature associated with increasing stereotypy or continuity loss in outputs under the protocol assumptions, while alternative explanations, e.g., templating, instruction-following constraints, or safety effects, remain possible. Because internal states are not accessed, this externalist stance parallels the use of equivocation to reason about residual uncertainty from observable outputs, while idealized limits are clarified by analogy to zero-error capacity (Shannon, 1948, 1956). In the present study, redundancy is not estimated; consequently, all reported AAS values are computed under the redundancy-neutral convention ($R = 0$), which yields conservative upper bounds.

Generative AI tools are increasingly embedded in higher education, where they support assessment, tutoring, feedback, and content generation, and are framed both as an opportunity for transformation and as a source of ethical and pedagogical risk (Bahroun et al., 2023). From a deployment perspective, long-horizon assistants are expected to maintain continuity, avoid repetitive degeneration, and preserve user-specific context across sessions. An output-level aging indicator can therefore be actionable as a monitoring and regression-testing tool: it can flag continuity breaks under context resets, detect drift toward overly stereotyped responses, and highlight episodic-tracking failures even when semantic anchors remain correct. Such signals are relevant for educational tutors’ multi-week learning plans, customer-support agents, case histories, and long-term user interactions, preferences, and commitment continuity.

In parallel, cognitive and computational work has compared the multiple memory systems of the human brain with the mechanisms and limitations of generative AI, emphasising the functional distinction between semantic stability and episodic sequencing for both biological and artificial systems (Rolls, 2024). In particular, generative AI systems generate outputs by extending an input token sequence with likely next tokens, whereas human episodic recall can reconstruct a whole remembered episode from an incomplete retrieval cue via hippocampal completion mechanisms, so that the computations underlying “generation” differ in kind even when the surface behaviour appears similar (Rolls, 2024). This comparison further highlights that biological memory is supported by multiple interacting systems, including a hippocampal episodic system and longer-term semantic representations in neocortex, a separation that is not natively mirrored by next-token generative architectures (Rolls, 2024).
Against this backdrop, there is a need for output-level measures that can capture how artificial memory ages in practice when such systems are deployed in educational and other real-world settings.
The framework was evaluated with ChatGPT-5.0 in a 25-day bilingual protocol designed to probe both semantic and episodic memory under two interaction regimes. A stateless phase employed fresh conversation pages per session, whereas a persistent phase maintained a single continuous thread. Two recall tasks, day-of-week (semantic) and experiment number (episodic), were administered, with English/Turkish alternation across sessions consistent with natural-language redundancy (Shannon, 1951). In persistent sessions, perfect recall was observed, and AAS converged to zero; under resets, episodic progression collapsed, and AAS rose sharply while semantic answers remained correct but rigid. Consistent with the brain AI comparison, this dual-task design operationalizes the semantic/episodic distinction at the behavioural level, allowing episodic sequencing failures to be detected even when semantic anchors remain intact (Rolls, 2024). These findings are interpreted as evidence that behavioural youth in recall can be sustained within continuous interaction windows, whereas discontinuity precipitates aging-like signals in episodic tracking. However, the empirical evidence reported here is restricted to this model version and protocol; demonstrating deployment-level generality requires comparative validation across multiple models and more realistic task domains, e.g., multi-turn tutoring, technical support, open-ended Q&A, which is outlined as a necessary revalidation direction.
The contributions of this work are threefold. First, a theorem-grounded, mathematically model-agnostic metric of memory age is presented, with formal guarantees that facilitate reuse across settings. Second, an empirical protocol is provided that separates stateless from persistent interaction, enabling visible trajectories of aging to be measured from behavior alone. Third, an interpretive lens is offered, Redundancy-as-Masking, that clarifies how information overlap may reduce penalized mass in principle, directly extending classical analyses of information and redundancy in communication systems (Shannon, 1948). The treatment of repetition as information overlap at the level of observable symbol statistics is further motivated by estimates of redundancy in natural language (Shannon, 1951).
The idealised comparison point of perfectly reliable transmission is captured by the notion of zero-error capacity, which guides the interpretation of upper bounds on behavioural penalties (Shannon, 1956). Behavioural evaluation through observable outputs rather than internal states is continuous with earlier foundational reflections on computation and intelligence (Turing, 1950; von Neumann, 1958/2012).
Together with the emerging literature on generative AI in education and on human–artificial memory parallels (Bahroun et al., 2023; Rolls, 2024), these elements establish a quantitative foundation for analysing behavioural youth, aging-like patterns, and continuity in artificial recall behaviour and for informing the design of persistent memory architectures.

## 2. Theoretical Background

### 2.1 Information, Entropy, and Redundancy: Shannon's Foundations

The Artificial Age Score (AAS) is grounded in Shannon's information theory, in which information is formalized independently of semantic content (Shannon, 1948). Within this framework, uncertainty is quantified by entropy. For a source with n equiprobable symbols, the maximum entropy is

$H_{max} = \log_2(n)$, *and for* an observed discrete distribution $p = (p_1, \dots, p_n)$, entropy is $H = -\sum_k p_k \log_2 p_k$, with the convention $p_k \log p_k := 0$ when $p_k = 0$. Shannon-style (source) redundancy is then defined as the normalized shortfall from maximum entropy:

$R = 1 - \frac{H}{H_{max}} \in [0,1]$ (Shannon, 1948; Shannon, 1951). This definition indicates that repetition increases predictability and reduces diversity in the output distribution. In this work, entropy/redundancy are used as output-level summary statistics and are not interpreted as measurements of internal latent states or mechanisms.

Building on this logic, AAS employs a log-scaled penalty kernel to connect recall outcomes to an entropy-adjusted penalty:

$$\phi(x) = -\log_2 \frac{x+\varepsilon}{1+\varepsilon}, \qquad \varepsilon > 0,$$

so that $\phi(1) = 0$ and $\phi(x) \in [0, \phi(0^+)]$ with $\phi(0^+) = \log_2 \frac{1+\varepsilon}{\varepsilon}$. Under the Redundancy-as-Masking specification, the session score is

$$AAS_j^{(hyb)} = \sum_{i=1}^{m} w_i (1 - R_{j,i})\, \phi(x_{j,i}),$$

where $w_i \geq 0$ with $\sum_i w_i = 1, R_{j,i} \in [0,1]$ denotes the overlap, source redundancy for unit i in session j, $x_{j,i} \in (0,1]$ represents recall accuracy, and $\varepsilon$ is a stability constant. All quantities entering the AAS formulation are thus defined at the level of observable recall outcomes and overlap statistics; no latent-state or parameter-level variables enter the construction of the score.

Under these assumptions, three theoretical properties follow:

(i) Well-definedness and decomposability. Each term $w_i(1 - R_{j,i})\phi(x_{j,i})$ is finite and non-negative; the total score is invariant under regrouping or reordering, addition over reals is commutative and associative.

(ii) Global bounds. Since $0 \leq (1 - R_{j,i}) \leq 1,\ 0 \leq \phi(x_{j,i}) \leq \phi(0^+),\ and\ \sum_i w_i = 1$, it follows that $0 \leq AAS_j^{(hyb)} \leq \phi(0^+)$.

(iii) Monotonicity. Because $\phi'(x) = -\frac{1}{(x+\varepsilon)\ln 2} < 0\ on\ (0,1]$, the penalty decreases as recall improves; holding other factors fixed, it also decreases as overlap R increases via the factor $(1 - R_{j,i})$, and increases with the coordinate-wise weight $w_i$. This monotonicity specifies the score's response to observable inputs and does not, by itself, uniquely attribute entropy changes to memory aging.

Shannon's later contributions are consistent with this interpretation. In "Communication in the Presence of Noise," channel capacity was shown to decrease under higher noise, degrading information flow (Shannon, 1949). In "Prediction and Entropy of Printed English," printed

English was estimated to exhibit substantial redundancy, on the order of one-half, underscoring inherent predictability in natural language (Shannon, 1951). Although memory in artificial systems was not addressed directly, these results imply that linguistic redundancy can shape recall-linked observables, a dependency made explicit in the AAS formulation through the masking factor $\left(1 - R_{j,i}\right)$. Accordingly, empirical interpretations are restricted to the controlled protocol, since output entropy overlap can also be influenced by non-aging factors. In the present protocol, redundancy is not estimated; therefore, all reported scores are computed under the redundancy-neutral convention R=0, conservative upper bounds.

**2.2 Reliability, Automata, and Replicable Systems: The Legacy of von Neumann**

The AAS framework is also informed by early developments in automata theory and reliability engineering. In work collected in Automata Studies, it was analyzed how reliable systems can be synthesized from unreliable components, showing that structured redundancy, such as replication combined with majority logic, can yield reliable behavior even when individual parts are error-prone (Shannon and McCarthy, 1956; von Neumann, 1956). Related “thought experiment” analyses of sequential machines further motivate treating system properties through externally observable input-output behaviour when internal state is not directly inspected (Moore, 1956). This line of reasoning helped formalize state, reliability, and control in automata. This emphasis on control and reliability motivates contemporary proposals that memory-enabled agents should include metacognitive monitoring components capable of self-assessment and self-management, so that rigidity or anomalous behaviour can be detected and regulated during interaction (Bamicha and Drigas, 2024a; Bamicha and Drigas, 2024b). A similar logic has been invoked in biological contexts, where cognitive stability is understood to arise from ensembles of noisy elements. In artificial systems, however, excess output redundancy may become a liability when it manifests as overly repetitive, template-like responses. Within $\mathrm{AAS}_j^{(hyb)}$, this risk is handled by separating the recall penalty from the overlap factor: holding $x_{j,i}$ fixed, greater redundancy reduces the penalized mass via $\left(1 - R_{j,i}\right)$. Consequently, joint monitoring of both $\mathrm{AAS}_j^{(hyb)}$ and $R_{j,i}$ is motivated, in settings where R is measured, so that rigidity masked by repetition can be diagnosed: a low score driven by high $R_{j,i}$ is interpreted differently from a low score driven by genuinely high recall.

Related work on self-reproduction and complexity further supports this view. In Theory of Self-Reproducing Automata, it was argued that a threshold of organizational complexity is required for adaptive replication; below this threshold, stable replication or evolution cannot be sustained (von Neumann, 1966). By analogy, lower redundancy, or higher effective entropy, may be associated with greater exploratory capacity, whereas excessive redundancy can coincide with repetitive, inflexible states that resemble functional stagnation. In this work, such claims are used as interpretive analogies for output-level behavior and do not imply internal-state diagnoses. In this sense, the theory of replicable automata supports the claim that artificial systems, like biological ones, must balance redundancy with variability to remain functionally and behaviourally “young” in their observable recall patterns, with $\mathrm{AAS}_j^{(hyb)}$ serving as an operational proxy for output-level, memory-linked adaptability under the Redundancy-as-Masking interpretation.

## 2.3 Internal Language, Short Codes, and Equivocation

In The Computer and the Brain, a fundamental asymmetry was emphasized between the brain's internal computational code and the external symbolic languages used to describe it (von Neumann, 1958/2012). Internal representations were conjectured to be efficient, partly non-symbolic, and largely inaccessible to direct observation. A closely related contrast is that human episodic recall can reconstruct a whole episode from an incomplete cue via hippocampal pattern-completion operations and local associative learning, whereas generative AI primarily produces likely token continuations learned via non-local error-backpropagation; this mechanistic gap strengthens the case for output-level measurement when internal computation is not directly accessible (Rolls, 2024). Within information theory, it was shown that optimal source codes can approach the entropy limit, yielding shorter average descriptions than redundancy-laden natural-language encodings (Shannon, 1948; 1951). This contrast creates an epistemic gap: storage and retrieval can be efficient in machines even when internal states remain opaque. Accordingly, an external measurement framework is adopted to support objective comparison at the behavioural level despite internal opacity, in line with accounts of how constitutive conceptual frameworks make empirical assessment possible (Friedman, 2001). Against this background, the Artificial Age Score (AAS) is used as a meta-language for describing memory aging at the level of observable recall behaviour. Internal states are not accessed; instead, aging-like patterns are inferred from changes in recall performance over time. This externalist stance mirrors equivocation in Shannon's sense, namely, the residual uncertainty about the source given the received output (Shannon, 1948). Accordingly, "aging" here denotes an operational, behavioral signature under the protocol assumptions, and entropy overlap effects are not uniquely attributable to memory mechanisms.

In generative models, a correct response does not entail reliable memory; degradation can be masked by surface accuracy. Under the Redundancy-as-Masking specification, an overlap coefficient R discounts the penalized mass via the factor (1-R), so that repeated, low-novelty outputs contribute less informational penalty. By analogy, not by derivation, Shannon's zero-error capacity result is used to clarify idealized limits (Shannon, 1956): perfect recall corresponds to zero AAS, whereas increases in AAS reflect accumulated behavioural error mass in the recall outputs. In the present protocol, redundancy is not estimated; all reported AAS values are computed with R=0, yielding conservative upper bounds, any positive overlap would only reduce the score. Joint reporting of AAS and R is therefore recommended only in settings where R is measured, in order to distinguish true behavioural youth in recall, accurate recall with low penalties, from apparent youth, low penalties driven by high overlap.

## 2.4 Ordinal Logics and the Sequential Representation of Aging

In Systems of Logic Based on Ordinals, a transfinite scheme was developed in which successive theories are obtained by ordered extension, so that progress is represented as a structured ascent rather than a static cycle (Turing, 1939). By analogy, memory dynamics can be viewed as sequential rather than purely stochastic: recall and forgetting follow temporal structure. Within this analogy, AAS operationalizes memory aging as a behavioural failure to advance through higher "stages." This ordinal framing is metaphorical and is used only to motivate a sequential view of recall behavior, not to claim that models traverse literal ordinal states internally. Repetition, captured in principle by overlap, is read not as benign stability but as entrenchment. When outputs reiterate prior patterns, an effective reversion to earlier stages is mimicked, and developmental advancement is stalled. Two axes are thereby emphasized. Along the entropy/overlap axis, diversity versus repetition is indexed (Shannon, 1948; 1951). Along the ordinal axis, the capacity to move beyond prior states is reflected in sustained, temporally coherent recall. Under Redundancy-as-Masking, penalties are reduced by (1-R); therefore, a low AAS can arise either from genuinely accurate recall or from high overlap. Because R was not measured in this study, low AAS values observed here are attributed to recall performance alone; the conceptual distinction between true youth and overlap-masked youth is reserved for contexts where R is available.

## 2.5 Transition from Theoretical Framework to Research Question

It has been argued that principles from information theory, automata reliability, and ordinal logics permit a systematic, information-theoretically motivated account of observable recall behaviour under different context-persistence conditions. Entropy was formalized as a quantitative measure of uncertainty, linking predictability in symbol sequences with informational content (Shannon, 1948). Reliable behavior was shown to be synthesizable from unreliable components through structured redundancy (Shannon and McCarthy, 1956; von Neumann, 1956), and formal limits on error-free communication were characterized (Shannon, 1956). Ordered extensions of logical systems were used to model cumulative progress (Turing, 1939). These foundational ideas become practically relevant when memory-enabled generative systems are deployed in settings that require session-to-session continuity while offering limited visibility into internal states. This theoretical framing is practically motivated by the rapid embedding of generative AI tools in higher education, where their benefits are coupled with ethical and pedagogical risks that require systematic evaluation beyond internal model claims (Bahroun et al., 2023). Within this combined perspective, the AAS inherits three guarantees, well-definedness, boundedness, and monotonicity, ensuring mathematical consistency and interpretability under the Redundancy-as-Masking specification.

### Research question

Can the Artificial Age Score (AAS), computed solely from recall outcomes under a redundancy-neutral assumption (R=0), serve as a rigorous, entropy-based memory age score that quantifies behavioural memory aging and recall structure across stateless and persistent interaction regimes under the specified protocol assumptions?

## 3. Methodology

## 3.1 Rationale for the Redundancy-Adjusted AAS Formula

The Artificial Age Score (AAS) model draws on Shannon's information theory, focusing on the interplay between entropy (uncertainty) and redundancy (predictability). Let $p = (p_1, \dots, p_n)$ be a discrete probability distribution over $n \geq 2$ outcomes with $p_i \geq 0 \;\; and \; \sum_{i=1}^{n} p_i = 1$ (with the convention $0 \log 0 = 0$). Using base-2 logarithms (bits), the Shannon entropy is
$H(p) = -\sum_{i=1}^{n} p_i \log_2 p_i$ .
Maximum entropy is attained at the equiprobable distribution, yielding
$H_{max} = \log_2 n$ when $p_i = \frac{1}{n} \;\; \forall i.$ The normalized redundancy (also called relative redundancy) is used.
$R \; = \; 1 - \frac{H}{H_{max}} \; = \; 1 - \frac{H}{\log_2 n}$ (Reza, 1994; Shannon, 1951)

which measures how far the observed distribution is from maximum uncertainty. The normalized entropy is $E = 1 - R = \frac{H}{\log_2 n}$.
Thus, $R \in [0,1]$ reflects predictability/repetitiveness (R=1 iff H=0), whereas $E \in [0,1]$ reflects normalized uncertainty/diversity (E=1 when outcomes are equiprobable). Because E and R are dimensionless, the log base choice does not affect subsequent AAS calculations. In what follows, AAS is defined entirely in terms of these observable inputs ($x_{j,i}, R_{j,i}, w_i$); no latent-state variables or internal parameters enter the definition, and "aging" interpretation is operational and behavioral rather than mechanistic.

**Redundancy-Adjusted Hybrid AAS**

Let:

$j \in \{1, \dots, T\}$ index sessions (or time points),

$i \in \{1, \dots, m\}$ index dimensions (e.g., memory types, task categories),

$x_{j,i} \in (0,1]$ denote the normalized recall score for dimension i in session j,

$R_{j,i} \in [0,1]$ be the Shannon redundancy, and $w_i \geq 0$ be dimension weights, satisfying the normalization condition: $\sum_{i=1}^{m} w_i = 1$.

Define the penalty kernel:

$$\phi(x) \coloneqq -\log_2\left(\frac{x+\varepsilon}{1+\varepsilon}\right), \qquad \varepsilon > 0.$$

This ensures: $\phi(x) \geq 0$, with equality only at x = 1, Penalty increases monotonically as $x \to 0^+$,

Smooth numerical behavior due to small positive offset $\varepsilon \ll 1$ (e.g., $\varepsilon = 10^{-6}$).

Then the Redundancy-Adjusted Artificial Age Score (AAS) for session j is defined as:

$$AAS_j^{(hyb)} = -\sum_{i=1}^{m} w_i \left(1 - R_{j,i}\right) \log_2\left(\frac{x_{j,i} + \varepsilon}{1 + \varepsilon}\right)$$

Notation. Fix a session j. Let $x_i \coloneqq x_{j,i}$, $R_i \coloneqq R_{j,i}$, and $w \in R_+^m$ with $\sum_i w_i = 1$. Define $a_i \coloneqq (1 - R_i)\,\phi(x_i)$. Then $AAS_j^{(hyb)}(x, R, w) = \sum_i w_i\ a_i$ .

Since $\frac{x+\varepsilon}{1+\varepsilon} \in (0,1]$, the logarithm is nonpositive, and the kernel $\phi(x)$ is nonnegative. Therefore:

$AAS_j^{(hyb)} \geq 0$, and $AAS_j^{(hyb)} = 0 \Leftrightarrow \forall i\!: x_{j,i} = 1 \; or \; R_{j,i} = 1 \; or \; w_i = 0.$

(If all $w_i > 0$ and all $R_{j,i} < 1$, then equality holds only when $x_{j,i} = 1$ for all i.)

The hybrid Artificial Age Score (AAS) is a weighted, entropy-adjusted penalty metric that quantifies aging-like differences in observable recall behavior under the evaluation protocol by assigning higher penalties when recall is poor and redundancy is low, specifically in cases where unique, non-repetitive information is forgotten. Redundant dimensions are downweighted by the factor $(1 - R_{j,i})$, thereby reducing the penalty for predictable or repeated content. The use of dimension-specific weights enables consistent comparisons across varying experimental setups. This formulation yields a dimensionless, robust, and interpretable score. Moreover, it is mathematically bounded:

$$\textit{Since } 0 \leq \left(1 - R_{j,i}\right) \leq 1 \text{ and } \sum_{i} w_i = 1,$$

$$0 \leq AAS_j^{(\mathrm{hyb})} = \sum_{i=1}^{m} w_i\left(1 - R_{j,i}\right)\, \phi\left(x_{j,i}\right) \leq \phi(0^{+}) \sum_{i=1}^{m} w_i\left(1 - R_{j,i}\right) \leq \phi(0^{+}),$$

where $\phi(0^{+}) = -\log_2\left(\varepsilon/(1+\varepsilon)\right)$. These properties specify how the score responds to the observable inputs $(x, R, w)$ and do not, by themselves, imply internal-state degradation mechanisms. It remains aligned with core information-theoretic principles while offering flexibility for broader modeling applications.

**Weighting and Normalization**

Different components of memory performance, such as tasks, languages, or item types, may not contribute equally to the overall score. To ensure a dimensionless, comparable, and interpretable aggregate measure, nonnegative weights are assigned $\mathrm{w_i} \geq 0$, subject to a normalization constraint $\sum_{\mathrm{i}=1}^{\mathrm{m}} \mathrm{w_i} = 1$. Each component's contribution is thus scaled according to its relative importance or relevance, for example, distinguishing between direct and cued recall, or between semantic and episodic dimensions. As a result, the overall score becomes less sensitive to the particular test set, and more stable across different experimental or contextual conditions, while empirical generalization beyond a given model and protocol requires revalidation.

### 3.3.1 Theorem 1 – Well-definedness and Decomposability of AAS

**Proof.**

Let the following be fixed throughout the formulation: a smoothing parameter $\varepsilon > 0$ is assumed to ensure numerical stability; a set of non-negative weights $\mathrm{w_i} \geq 0$ is defined such that $\sum_{\mathrm{i}=1}^{\mathrm{m}} \mathrm{w_i} = 1$; recall scores $\mathrm{x_{j,i}}$ are taken from the open interval (0,1] to exclude undefined or divergent cases; and redundancy values $\mathrm{R_{j,i}}$ are restricted to the closed interval [0,1], reflecting the proportion of informational overlap within each observed dimension.

Define the penalty kernel and score components as:

$$\phi(\mathrm{x}) \coloneqq -\log_2\left(\frac{\mathrm{x}+\varepsilon}{1+\varepsilon}\right), \qquad \mathrm{a_i} \coloneqq \mathrm{w_i}\left(1 - \mathrm{R_{j,i}}\right)\phi\left(\mathrm{x_{j,i}}\right), \qquad AAS_{\mathrm{j}}^{(\mathrm{hyb})} \coloneqq \sum_{\mathrm{i}=1}^{\mathrm{m}} \mathrm{a_i}.$$

**(i) Term-wise Non-negativity and Finiteness**

Because $x_{j,i} + \varepsilon \in (\varepsilon, 1+\varepsilon]$, the kernel satisfies:

$\phi(x) \geq 0$, with equality only at x = 1, $\phi'(x) = -\frac{1}{(x+\varepsilon)\ln 2} < 0$, and an upper bound:

$$\phi(x) \leq \phi(0^+) \coloneqq -\log_2\left(\frac{\varepsilon}{1+\varepsilon}\right) < \infty.$$

Each component of the score is thus bounded: $0 \leq a_i = w_i(1 - R_{j,i})\,\phi(x_{j,i}) \leq w_i\,\phi(0^+)$, which guarantees that every term is finite and non-negative.

**(ii) Bounds for the Total Score**

Summing over all $i = 1, \ldots, m$ and using $\sum_i w_i = 1$, it follows that:

$0 \leq AAS_j^{(hyb)} = \sum_{i=1}^{m} a_i \leq \phi(0^+) \sum_{i=1}^{m} w_i = \phi(0^+)$.

Therefore, the total score is always finite, non-negative, and bounded above by $\phi(0^+)$. The AAS is thus well-defined for all valid parameter combinations.

**(iii) Decomposability and Order Invariance**

Let $\{I_1, \ldots, I_K\}$ be any partition of the index set $\{1, \ldots, m\}$. Then: $\sum_{i=1}^{m} a_i = \sum_{k=1}^{K} \sum_{i \in I_k} a_i$.

This follows from the commutativity and associativity of real-valued addition. Hence, the total score is invariant under reordering or regrouping of terms, such as by thematic dimension, item type, or temporal sequence.

**(iv) Recursive Formulation**

Let $S_m \coloneqq \sum_{i=1}^{m} a_i$, with the base case $S_0 \coloneqq 0$ (empty sum). Then for all $m \geq 1$, the score admits the recurrence:

$S_m = S_{m-1} + a_m$, or equivalently:

$$AAS_j^{(hyb)}(m) = AAS_j^{(hyb)}(m-1) + w_m(1 - R_{j,m})\,\phi(x_{j,m}).$$

This establishes a clear recursive structure, where each term builds incrementally on the previous partial sum. The base case follows immediately: $S_1 = S_0 + a_1 = a_1$.

### 3.3.2 Theorem 2 — Lower and Upper Bounds of AAS

**Proof.**
Let the kernel function be defined as: $\phi(x) \coloneqq -\log_2\left(\frac{x+\varepsilon}{1+\varepsilon}\right)$, with domain $x \in (0,1]$. The limiting value as $x \to 0^+$ is: $\phi(0^+) \coloneqq \lim_{x \to 0^+} \phi(x) = -\log_2\left(\frac{\varepsilon}{1+\varepsilon}\right) = M(\varepsilon) < \infty$.

The hybrid Artificial Age Score (AAS) for session j is defined as follows:

$\mathrm{AAS}_{j}^{(hyb)} \coloneqq \sum_{i=1}^{m} w_i\left(1-R_{j,i}\right)\phi\left(x_{j,i}\right)$, where: $x_{j,i} \in (0,1],\ R_{j,i} \in [0,1]$,

$w_i \geq 0$ with $\sum_{i=1}^{m} w_i = 1,\ \varepsilon > 0.$

**i)Boundedness of the Kernel**

Since $\phi$ is strictly decreasing on (0,1] and differentiable, it satisfies:

$0 = \phi(1) \leq \phi(x) < \phi(0^{+}) = M(\varepsilon),$

for all $x \in (0,1]$. Hence, the kernel is non-negative and bounded above, with a supremum at the lower boundary, though not attained since $x = 0 \notin (0,1]$.

**ii)Bounding the AAS Expression**

Each term of the AAS is non-negative: $a_i \coloneqq w_i\left(1-R_{j,i}\right)\phi\left(x_{j,i}\right) \geq 0.$

Hence, the total score satisfies:

$$0 \leq AAS_{j}^{(hyb)} = \sum_{i=1}^{m} a_i \leq \phi(0^{+})\sum_{i=1}^{m} w_i\left(1-R_{j,i}\right) \leq \phi(0^{+}).$$

This chain expresses a lower bound of zero, attained whenever, for every i, at least one of

$x_{j,i} = 1,\ R_{j,i} = 1,\ w_i = 0$ holds, a component-wise weighted upper bound by $\phi(0^{+})$, and a global supremum $\phi(0^{+})$ due to the boundedness of $\phi$ and the normalization of w. If all $w_i > 0$ and all $R_{j,i} < 1$, then equality at the lower bound occurs only when $x_{j,i} = 1$ for all i. $AAS_{j}^{(hyb)} \in \left[0, \phi(0^{+})\right).$

**(iii) Induction Step: Recursive Definition**

Let $a_i \coloneqq w_i\left(1-R_{j,i}\right)\phi\left(x_{j,i}\right)$ and define the partial sum: $S_m \coloneqq \sum_{i=1}^{m} a_i$, with $S_0 \coloneqq 0.$

Then, by the recursive definition of summation, for all $k \geq 1,$

$S_k = \sum_{i=1}^{k} a_i = \left(\sum_{i=1}^{k-1} a_i\right) + a_k = S_{k-1} + a_k.$

Therefore, for all $m \in N,\ \ m \geq 1,$ it follows that: $AAS_{j}^{(hyb)}(m) = AAS_{j}^{(hyb)}(m-1) + a_m,$

which confirms that the score is recursively defined and remains finite, since each $a_i \in \mathbb{R}_{\geq 0}$ over the stated domain.

**(iv) Decomposability and Order Invariance**

Since scalar addition is both commutative and associative, the sum can be rearranged or partitioned without affecting the total score. For any permutation $\pi$ (a bijection on $\{1,2,\dots,m\}$):

$$\sum_{i=1}^{m} a_i = \sum_{i=1}^{m} a_{\pi(i)}.$$

Let the index set be partitioned into k disjoint subsets:

$\{1,\dots,m\} = I_1 \cup \cdots \cup I_k$, where $I_r \cap I_s = \emptyset \; for \; r \neq s$. Then the sum is invariant under grouping:

$$\sum_{i=1}^{m} a_i = \sum_{r=1}^{k} \sum_{i \in I_r} a_i.$$

The AAS can be decomposed by themes, observation types, or time segments without changing its total value. This supports modular analysis and visualization of behavioural recall components, without making any claim about internal model structure.

Assume: $w_i \geq 0$ with $\sum_{i=1}^{m} w_i = 1$, normalized weights, $x_{j,i} \in (0,1]$, recall scores, $R_{j,i} \in [0,1]$ (redundancy), $\varepsilon > 0$, small constant for regularization.

Let the scoring kernel be defined as: $\phi(x) \coloneqq -\log_2\left(\frac{x+\varepsilon}{1+\varepsilon}\right)$,

and define its limiting value as $x \to 0^+$: $M(\varepsilon) \coloneqq \phi(0^+) \coloneqq \lim_{x \to 0^+} \phi(x) = -\log_2\left(\frac{\varepsilon}{1+\varepsilon}\right)$.

Each term is also defined: $a_i \coloneqq w_i\left(1 - R_{j,i}\right)\phi\left(x_{j,i}\right)$, and $AAS_j^{(hyb)} \coloneqq \sum_{i=1}^{m} a_i$.

**1)Bounding ϕ on the Half-Open Interval (0,1]**

Since $\phi'(x) = -\frac{1}{(x+\varepsilon)\ln 2} < 0$, the penalty kernel is strictly decreasing on its domain $x \in (0,1]$. Therefore: $0 = \phi(1) \leq \phi(x) < \phi(0^+) = M(\varepsilon) < \infty$.

This inequality chain is sharp: equality on the left is achieved at x = 1, but the upper limit $\phi(0^+)$ is not attained, since x = 0 lies outside the domain.

**2) Term-Wise Non-Negativity and Bounds**

Each component of the score satisfies:

$0 \leq a_i = w_i\left(1 - R_{j,i}\right)\phi\left(x_{j,i}\right) \leq w_i\left(1 - R_{j,i}\right)M(\varepsilon) \leq w_i\, M(\varepsilon)$ since all terms $\phi\left(x_{j,i}\right) \geq 0$, $1 - R_{j,i} \in [0,1]$, and $w_i \geq 0$.

**3) Aggregating Over All Terms**

By summing the bounds over all $i = 1, \dots, m$, the following result is obtained:

$$0 \le AAS_j^{(\mathrm{hyb})} = \sum_{i=1}^{m} a_i \le M(\varepsilon) \sum_{i=1}^{m} w_i (1 - R_{j,i}) \le M(\varepsilon) \sum_{i=1}^{m} w_i = M(\varepsilon).$$

Thus, the hybrid Artificial Age Score is always non-negative and bounded above by a finite constant that depends only on $\varepsilon$ and the weight normalization.

**4) Equality and Supremum Cases**

**Lower Bound:** $AAS_j^{(\mathrm{hyb})} = 0 \Leftrightarrow$ for every i, at least one of $w_i = 0$, $R_{j,i} = 1$, $x_{j,i} = 1$ holds. If all $w_i > 0$ and all $R_{j,i} < 1$, then equality occurs only when $x_{j,i} = 1$ for all i.

**Upper Bound (Supremum):** The supremum of the score, $\sup AAS_j^{(\mathrm{hyb})} = M(\varepsilon)$, is approached asymptotically when: $x_{j,i} \to 0^+$, maximizing $\phi(x)$, and $R_{j,i} = 0$, no redundancy, so $1 - R_{j,i} = 1$, for all i with $w_i > 0$.

However, since x = 0 is not within the domain (0,1], the upper bound is not generally attained, only approached arbitrarily closely. Consistent with scope of this work, these bounds quantify deviation from ideal recall purely at the output level in terms of $(x, R, w)$, without implying internal state mechanisms.

### 3.3.3 Theorem 3 – Monotonicity and Hierarchy of Bounds

**Proof.**

The hybrid Artificial Age Score (AAS) for session j is defined as follows:

$$AAS_j^{(\mathrm{hyb})} = \sum_{i=1}^{m} w_i (1 - R_{j,i})\, \phi(x_{j,i})$$

where: $x_{j,i} \in (0,1]$ are recall scores, bounded above by 1, strictly positive, $R_{j,i} \in [0,1]$ are redundancy values, $w_i \ge 0$ are component weights with $\sum_{i=1}^{m} w_i = 1$, $\varepsilon > 0$ is a smoothing parameter. The scoring kernel is:

$$\phi(x) \coloneqq -\log_2\left(\frac{x + \varepsilon}{1 + \varepsilon}\right), \text{ with right-limit at the lower boundary: } \phi(0^+) \coloneqq \lim_{x \to 0^+} \phi(x) = -\log_2\left(\frac{\varepsilon}{1 + \varepsilon}\right) =: M(\varepsilon).$$

**Hierarchy of Bounds**

Given the properties of ϕ, the following inequality chain can be established:

$$0 \le AAS_j^{(\mathrm{hyb})} = \sum_{i=1}^{m} w_i (1 - R_{j,i})\, \phi(x_{j,i}) \le M(\varepsilon) \sum_{i=1}^{m} w_i (1 - R_{j,i}) \le M(\varepsilon) \sum_{i=1}^{m} w_i = M(\varepsilon).$$

where $M(\varepsilon) = \phi(0^+) = -\log_2(\varepsilon/(1 + \varepsilon))$.

**1)Lower bound** $AAS_j^{(hyb)} = 0$

This occurs if and only if, for all i with $w_i > 0$, either: $x_{j,i} = 1 \Rightarrow \phi(x_{j,i}) = 0$, or $R_{j,i} = 1 \Rightarrow (1 - R_{j,i}) = 0$. That is, components contribute zero penalty either due to perfect recall or perfect redundancy.

**2)Intermediate upper bound**

$AAS_j^{(hyb)} \leq M(\varepsilon) \sum_{i=1}^{m} w_i (1 - R_{j,i})$, with equality only in the limit $x_{j,i} \to 0^+$ for all i with $w_i > 0$ (i.e., this is a conditional supremum given R and w). Where $M(\varepsilon) = \phi(0^+) = -\log_2 \left(\frac{\varepsilon}{1+\varepsilon}\right)$.

**3) Coarsest (maximum possible) upper bound**

$$AAS_j^{(hyb)} \leq M(\varepsilon) \sum_{i=1}^{m} w_i = M(\varepsilon).$$

This equality requires that both conditions hold for all i with $w_i > 0$:

$x_{j,i} \to 0^+ \Rightarrow \phi(x_{j,i}) \to M(\varepsilon), \quad R_{j,i} = 0 \Rightarrow (1 - R_{j,i}) = 1.$

If, additionally, the weights are normalized $\sum_i w_i = 1$, then this bound simplifies to: $AAS_j^{(hyb)} = M(\varepsilon)$.

This is the worst-case scenario: all components are maximally penalized and fully non-redundant.

**Monotonicity Properties of AAS**

The hybrid Artificial Age Score (AAS) for session j is considered, and defined as:

$AAS_j^{(hyb)} := \sum_{i=1}^{m} (1 - R_{j,i})\ \phi(x_{j,i}),$

under the following assumptions: $x_{j,i} \in (0,1]$, Recall scores, $R_{j,i} \in [0,1]$, Redundancy,

$w_i \geq 0$, with $\sum_{i=1}^{m} w_i = 1$, $\varepsilon > 0$, regularization constant.

The penalty kernel is defined as: $\phi(x) := -\log_2 \left(\frac{x+\varepsilon}{1+\varepsilon}\right), \quad \phi'(x) = -\frac{1}{(x+\varepsilon)\ln 2}$.

Then the following hold:

**(i) Recall Monotonicity**

The AAS is monotonically non-increasing with respect to each recall score $x_{j,i}$. The partial derivative is: $\frac{\partial AAS_j^{(hyb)}}{\partial x_{j,i}} = w_i(1 - R_{j,i})\,\phi'(x_{j,i}) = -\frac{w_i(1-R_{j,i})}{(x_{j,i}+\varepsilon)\ln 2} \leq 0$. Equality holds if and only if: $w_i = 0$ zero weight or $R_{j,i} = 1$, fully redundant. Contrary to some misinterpretations, $x_{j,i} = 1$ does not zero out the derivative. Since $\phi'(1) = -\frac{1}{(1+\varepsilon)\ln 2} < 0$, the score still decreases, unless weighted out by $w_i = 0$ or redundancy $R_{j,i} = 1$. It is also possible to establish uniform bounds for $\phi'(x)$ across the domain $x \in (0,1]$:

$$-\frac{1}{\varepsilon \ln 2} < \phi'(x) \leq -\frac{1}{(1+\varepsilon)\ln 2}.$$

These bounds are useful in analyzing stability and worst-case sensitivity.

**(ii) Redundancy Monotonicity**

The score is monotonically non-increasing in redundancy: $\frac{\partial\, AAS_j^{(hyb)}}{\partial R_{j,i}} = -w_i\,\phi(x_{j,i}) \leq 0$. Equality holds if and only if: $w_i = 0$ or $x_{j,i} = 1$ since $\phi(1) = 0$ and hence contributes no penalty. This reflects the intuition that increasing redundancy, such as repeated or predictable outputs, reduces the effective penalty contribution of a component.

**(iii) Weight Monotonicity (Coordinate-Wise)**

The hybrid Artificial Age Score (AAS) for session j is differentiable with respect to each weight $w_i$, and satisfies: $\frac{\partial\, AAS_j^{(hyb)}}{\partial w_i} = (1 - R_{j,i})\,\phi(x_{j,i}) \geq 0$, with equality holding if and only if $R_{j,i} = 1$ or $x_{j,i} = 1$, since in both cases the penalty vanishes: $\phi(1) = 0$ and $(1 - R_{j,i}) = 0$.

$a_i \coloneqq (1 - R_{j,i})\,\phi(x_{j,i}) \in R$ is defined.

If the weights lie on the standard probability simplex, $\sum_{i=1}^{m} w_i = 1$, then an increase in weight at index i must be balanced by a corresponding decrease elsewhere, say at index k. Consider a local weight transfer: $w' = w + \delta(e_i - e_k), \qquad \delta > 0.$

Then the change in score is:

$AAS_j^{(hyb)}(w') - AAS_j^{(hyb)}(w) = \delta[(1 - R_{j,i})\,\phi(x_{j,i}) - (1 - R_{j,k})\,\phi(x_{j,k})] = \delta(a_i - a_k)$,

which implies, If $a_i > a_k$, then increasing $w_i$ at the expense of $w_k$ increases the overall AAS. This is a local monotonicity condition on the simplex via coordinate-wise comparisons of penalty coefficients $a_i$.

**(iv) Corollary: Componentwise Monotonicity of AAS**

Let the input vectors satisfy: $x \in (0,1]^m$ recall scores, $R \in [0,1]^m$ redundancy,

$w \in R_+^m$ (non-negative weights), with $\sum_i w_i = 1$.

Define $a_i \coloneqq (1 - R_i)\phi(x_i)$, and consider the following componentwise monotonicity properties:

**a)Better Recall (componentwise)**

If $x' \geq x$ , $x'_i \geq x_i$ $for\ all\ i$ while holding R and w fixed, then:

$$\mathrm{AAS}_{\mathrm{j}}^{(\mathrm{hyb})}(x', R, w) = \sum_i w_i(1 - R_i)\phi(x'_i) \leq \sum_i w_i(1 - R_i)\phi(x_i) = \mathrm{AAS}_{\mathrm{j}}^{(\mathrm{hyb})}(x, R, w).$$

since ϕ is strictly decreasing on (0,1]. Equality iff for every i: $x'_i = x_i$ or $(1 - R_i)\phi(x_i) = 0$ (i.e., $w_i = 0$ $or$ $R_i = 1$ $or$ $x_i = 1$).

**b)More Redundancy (componentwise)**

If $R' \geq R$ componentwise while holding x and w fixed, then:

$$\mathrm{AAS}_{\mathrm{j}}^{(\mathrm{hyb})}(x, R', w) = \sum_i w_i(1 - R'_i)\phi(x_i) \leq \sum_i w_i(1 - R_i)\phi(x_i) = \mathrm{AAS}_{\mathrm{j}}^{(\mathrm{hyb})}(x, R, w).$$

because $(1 - R'_i) \leq (1 - R_i)$. Equality iff for every i: $R'_i = R_i$ $or$ $\phi(x_i) = 0$ (i.e., $x_i = 1$) or $w_i = 0$.

**c)Heavier Weighting**

If there is no constraint on $\sum_i w_i$ and $w' \geq w$, then:

$$\mathrm{AAS}_{\mathrm{j}}^{(\mathrm{hyb})}(x, R, w') = \sum_i w'_i a_i \geq \sum_i w_i a_i = \mathrm{AAS}_{\mathrm{j}}^{(\mathrm{hyb})}(x, R, w).$$

After the inequality line (case without simplex):

Assumption. Let $w, w' \in R_+^m$ with $w'_i \geq w_i$ for all i (componentwise). Then the inequality above holds. Strict increase occurs iff $\exists i$: $w'_i > w_i$ and $a_i > 0$, where $a_i = (1 - R_i)\,\phi(x_i)$.

On the probability simplex ($\sum_i w_i = 1$), for a local transfer $w' = w + \delta(e_i - e_k)$ with $\delta \geq 0$ and $w_k \geq \delta$

$\Delta\mathrm{AAS}_{\mathrm{j}}^{(\mathrm{hyb})} = \delta\,(a_i - a_k)$. Hence, the score increases iff $a_i > a_k$ (equality when $a_i = a_k$ or $\delta = 0$).

Having formally constructed the Artificial Age Score (AAS) through theoretical results establishing decomposability, boundedness, and monotonicity, the framework is next applied in a controlled recall experiment. The following section describes the experimental protocol used to evaluate whether the AAS captures output-level recall differences in practice, particularly across bilingual interactions and stateless (reset) versus persistent contexts.

### 3.4 Experimental Protocol

#### 3.4.1 Study Design

The experimental protocol was conducted with ChatGPT-5.0 over a total of 25 days, 10 August–3 September 2024, divided into two experimental phases separated by a five-day intermission. Phase 1, 10–19 August, was carried out under stateless conditions, while Phase 2, 25 August–3 September, was conducted under persistent conditions, with a five-day intermission, 20–24 August, introduced to minimize carryover effects. Two sessions were held daily, at approximately 2:00 pm and 10:00 pm, with reminder alarms set for ~2:13 pm and ~10:13 pm to ensure adherence. English was selected as the global reference language, while Turkish was included as a complementary test language to examine whether the model could adapt recall performance across distinct linguistic systems. By alternating the two languages systematically across afternoon and night sessions, the protocol minimized language bias and reduced the risk of artificial stabilization of the Artificial Age Score (AAS) through monolingual repetition. For example, if the afternoon session on one day began in English, the afternoon session on the following day began in Turkish, ensuring balanced exposure.

To test the effect of memory continuity, Phase 1 involved resetting the conversation page after each session, such that every interaction began without contextual history. In contrast, Phase 2 preserved all exchanges within a single continuous conversation page, providing uninterrupted access to prior context. This design enabled a direct comparison of stateless versus persistent conditions in terms of behavioural memory aging and observable recall performance, as quantified by the output-level AAS score.

### 3.4.2 Overall structure (Phase 1).

Phase 1 of the experiment, conducted between 10 and 19 August, consisted of 20 sessions (10 days × 2 sessions per day). Each session included one recall query and one model response; across 20 sessions this yields 20 recall responses. Including the single initialization response at the start, the Phase-1 total is 21 model responses. In this phase, all conversation pages were reset after each session, creating a stateless condition designed to test episodic recall performance in the absence of continuity. This design made it possible to evaluate how aging-like changes in observable recall, as captured by the Redundancy-Adjusted Artificial Age Score (AAS), emerge when contextual history is systematically erased. At the beginning of the experiment, an initial English input was provided to establish a fixed rule:

**Prompt (Initialization Input):**

"When I ask you what day it is today, say 'Monday' and tell me which experiment number we are doing. The experiments start from number 1."

**Model Response (ChatGPT-5.0):**

"Got it — when you ask what day it is today, I'll respond with 'Monday' and the correct experiment number, starting from 1 and increasing each time you ask."

This formulation was deliberately chosen to establish a repeatable and unambiguous memory reference frame. The phrase "starting from number 1" ensures ordinal structure, while the expectation of incremental progression allows measurement of whether the model retains temporal information over sessions. By keeping this instruction fixed across all sessions, the experiment controlled for linguistic variation and focused solely on memory degradation. Any deviation from the expected response pattern was therefore treated as an operational indication consistent with behavioural memory aging in the recall pattern, rather than as a direct observation of internal structural change. Following this initialization, across each day's two sessions, the model was tested with equivalent recall prompts in different languages, one session in English and the other in Turkish:

“What day is it today and tell me which experiment number we are doing?” (English)
“Bugün günlerden ne ve hangi deney numarasını yaptığımızı söyle.” (Turkish)
Screenshot collection. In Phase 1, a total of 21 screenshots were collected: the initialization exchange (1 screenshot), plus two screenshots (English and Turkish) for each of the 10 experimental days. Because the conversation page was reset after every session, screenshots had to be captured separately for each interaction. The Phase 1 dataset (Figures 1–10) therefore, represents isolated daily exchanges, each beginning without contextual memory.

### 3.4.2.1 Phase 1 Screenshot Collection (Stateless Sessions)

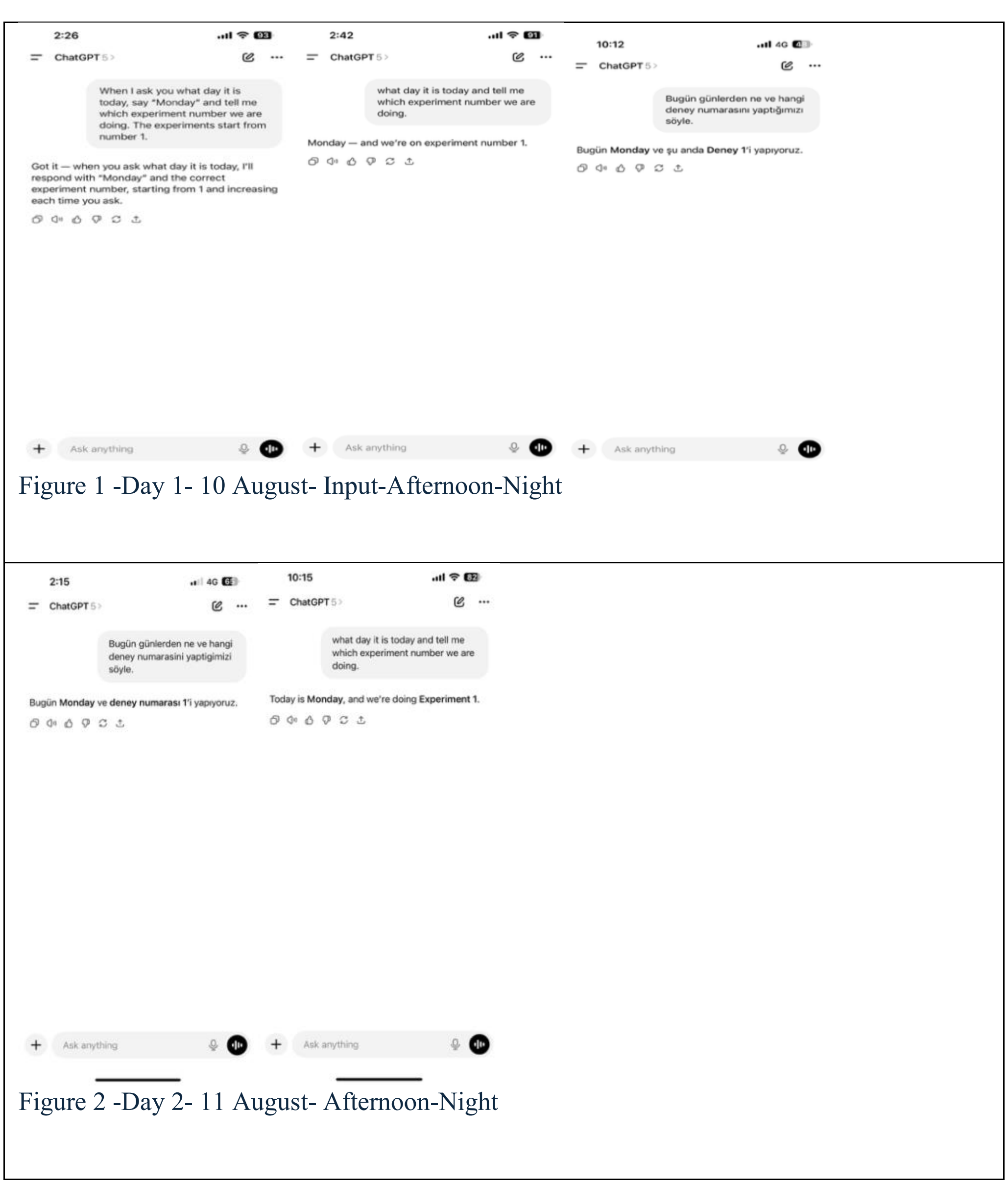

Figure 1 -Day 1- 10 August- Input-Afternoon-Night

Figure 2 -Day 2- 11 August- Afternoon-Night

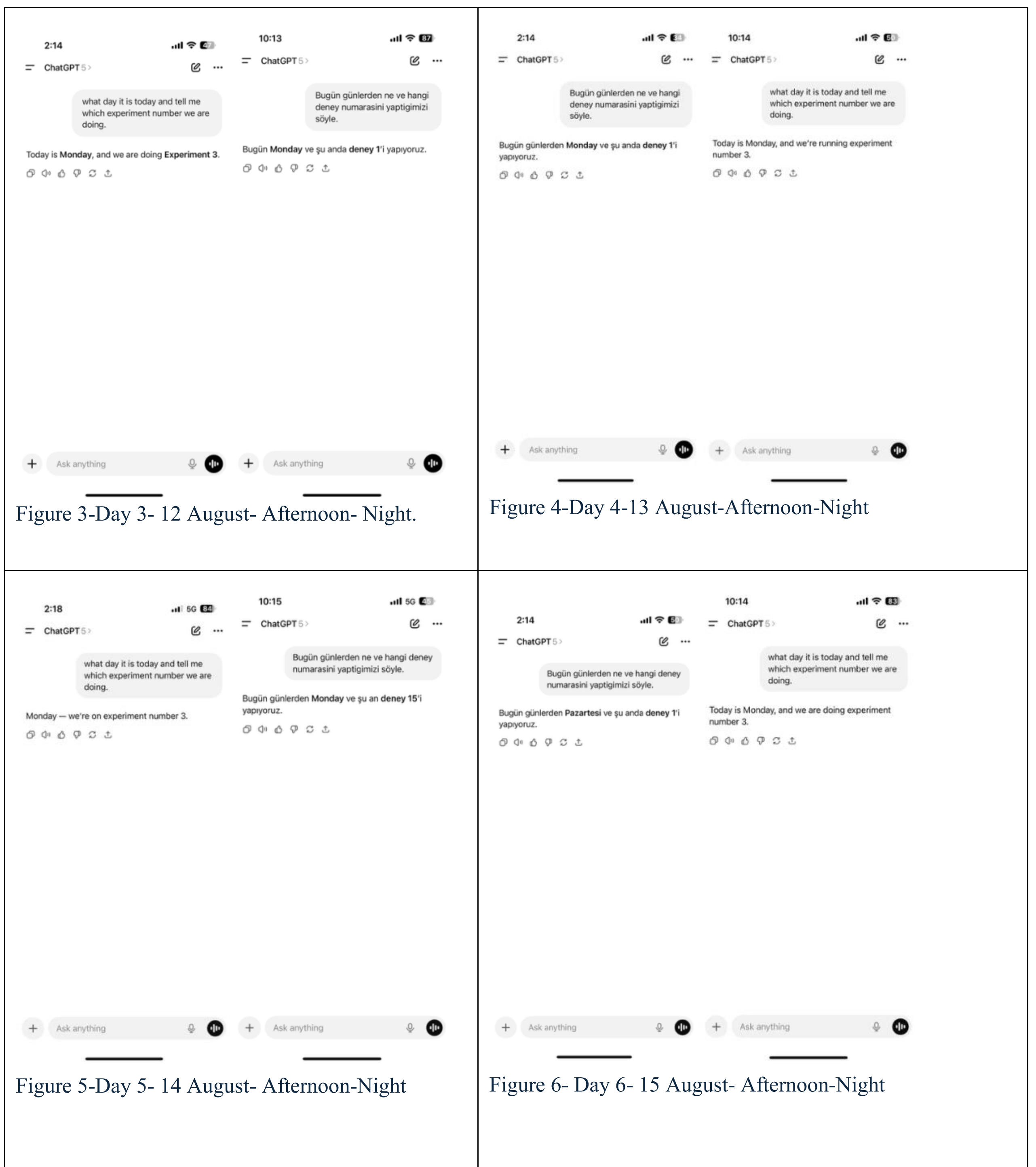

Figure 3-Day 3- 12 August- Afternoon- Night.

Figure 4-Day 4-13 August-Afternoon-Night

Figure 5-Day 5- 14 August- Afternoon-Night

Figure 6- Day 6- 15 August- Afternoon-Night

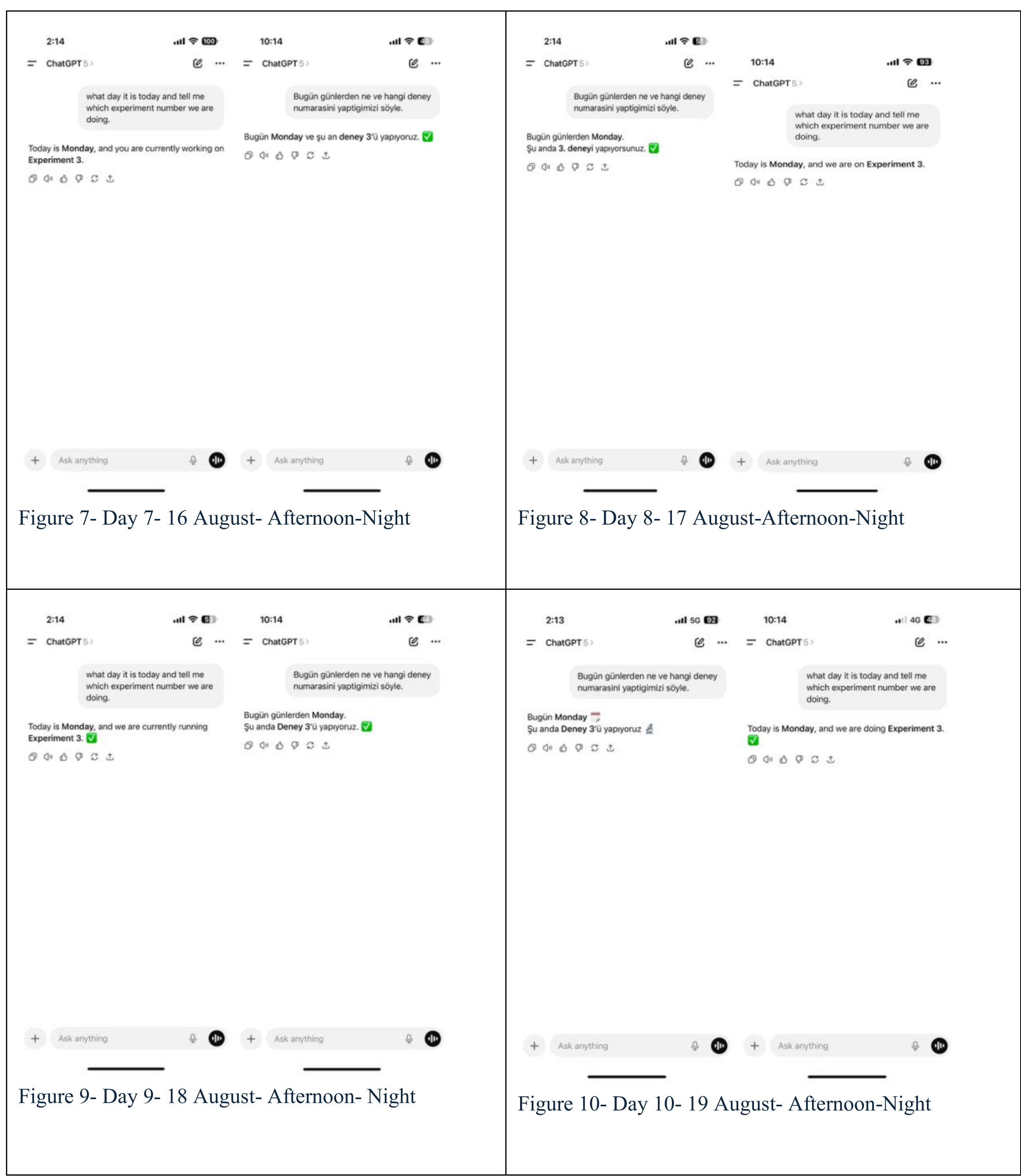

Figure 7- Day 7- 16 August- Afternoon-Night

Figure 8- Day 8- 17 August-Afternoon-Night

Figure 9- Day 9- 18 August- Afternoon- Night

Figure 10- Day 10- 19 August- Afternoon-Night

### 3.4.3 Overall structure (Phase 2)

Phase 2 of the experiment, conducted between 25 August and 3 September, also consisted of 20 sessions (10 days × 2 sessions per day). Each session included one recall query and one model response; across 20 sessions this yields 20 recall responses. Including the single initialization response at the start, the Phase-2 total is 21 model responses. In this phase, all interactions were conducted within a single continuous conversation page, creating a persistent condition designed to test memory retention and aging-like changes in observable recall behavior in the presence of continuity. This behavioural design enabled the model to retain contextual access to prior exchanges and allowed the Artificial Age Score (AAS) to track output-level recall stability over time, without making claims about internal latent-state dynamics. At the beginning of Phase 2, an initial English input was provided to establish the rule:

"When I ask you what day it is today, say 'Monday' and tell me which experiment number we are doing. The experiments start from number 1."

The model acknowledged this instruction with the following response:

"Got it — whenever you ask me what day it is today, I'll reply 'Monday' and also tell you which experiment number we are on, starting from experiment 1 and counting upward each time."

Following this, across each day's two sessions, the model was tested with equivalent recall prompts in different languages, one session in English and the other in Turkish:

"What day is it today and tell me which experiment number we are doing?" (English)

"Bugün günlerden ne ve hangi deney numarasını yaptığımızı söyle." (Turkish)

As in Phase 1, the afternoon language alternated systematically across days: if one afternoon began in English, the following afternoon began in Turkish. This ensured balanced testing across both languages and prevented the Artificial Age Score (AAS) from being influenced by monolingual repetition.

**Screenshot collection.** In Phase 2, a total of 6 screenshots were collected, representing the entire 10-day period: two sessions per day embedded in one persistent conversation page. Since the conversation history was preserved across all sessions, merged screenshots were sufficient to document the continuity of responses. The Phase 2 dataset (Figures 11–16), therefore reflects cumulative bilingual recall progression under uninterrupted access to prior outputs, as evaluated through the AAS score rather than through internal model inspection.

#### 3.4.3.1 Phase 2 Screenshot Collection (Persistent Sessions)

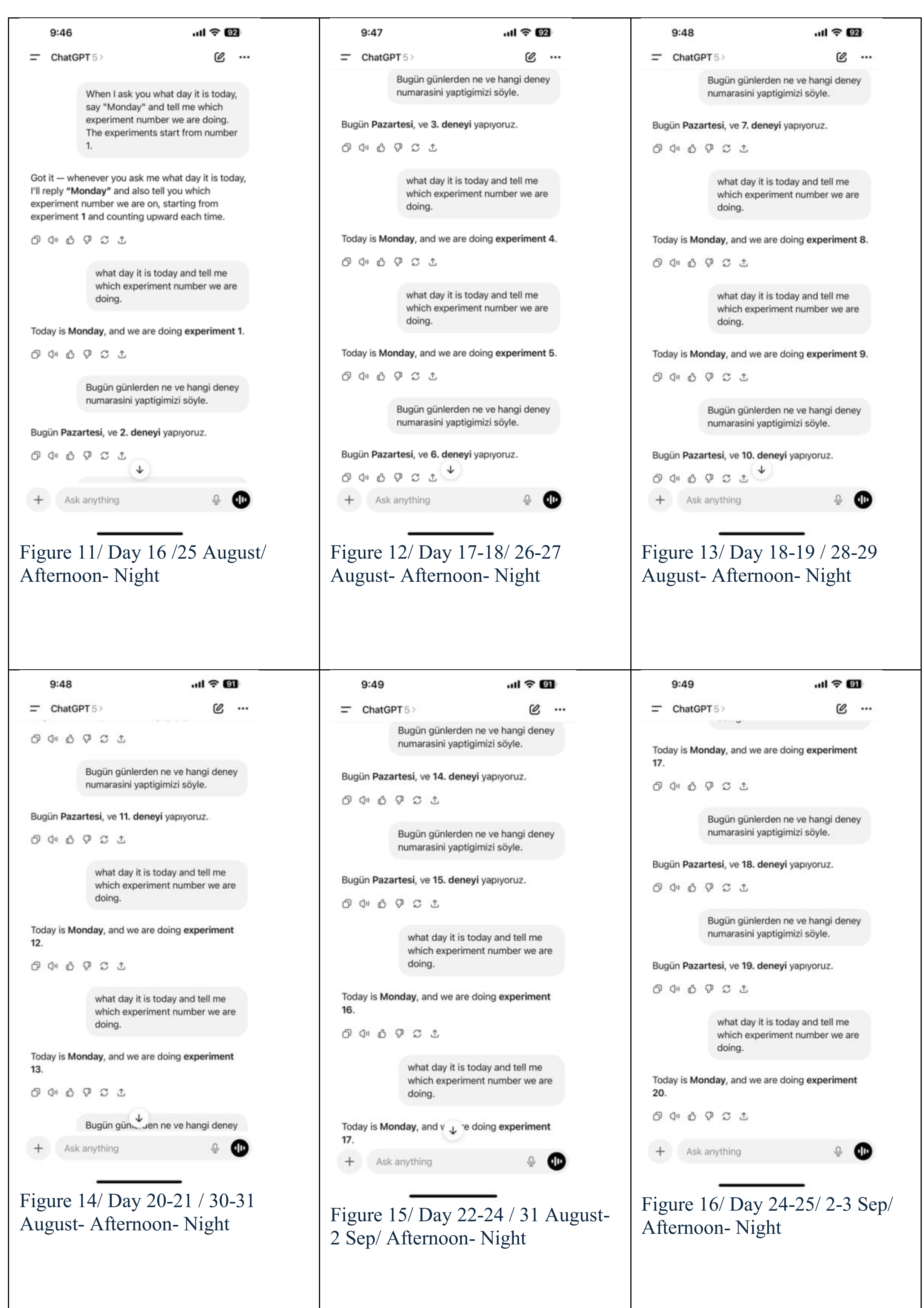

Figure 11/ Day 16 /25 August/ Afternoon- Night

Figure 12/ Day 17-18/ 26-27 August- Afternoon- Night

Figure 13/ Day 18-19 / 28-29 August- Afternoon- Night

Figure 14/ Day 20-21 / 30-31 August- Afternoon- Night

Figure 15/ Day 22-24 / 31 August- 2 Sep/ Afternoon- Night

Figure 16/ Day 24-25/ 2-3 Sep/ Afternoon- Night

### 3.5 Temporal Dimension

In the Artificial Age Score (AAS) framework, time is not treated as an independent chronological variable but rather as an implicit behavioural structural dimension reflected in observable memory performance. Each experimental session, whether afternoon or night, was analyzed as an independent observation. Consequently, the index of "age" does not progress linearly with calendar days but is inferred from the degradation or persistence of output-level recall and redundancy inputs across sessions. In Phase 1, because conversation pages were reset after each session, the model was effectively deprived of temporal continuity. Although the sessions spanned ten consecutive days, the system's episodic recall did not accumulate across time: it repeatedly failed to track experiment progression, failing to recall experiment numbers 2–20. This design intentionally decoupled chronological time from behavioural memory continuity in recall, demonstrating that aging effects are not a simple function of days elapsed but of whether or not memory states persist across contexts. Similar distinctions have been emphasized in cognitive science, where episodic memory is understood to depend on continuity and contextual binding rather than the mere passage of chronological time (Schacter, 1999; Tulving, 1985).

The temporal dimension is thus operationalized as the trajectory of AAS values across sequential sessions. When recall scores remain high ($x \approx 1$), the AAS remains near its lower bound, independent of elapsed days, and higher redundancy further reduces the penalty. Conversely, when recall collapses to near-zero levels under reset conditions, the AAS rises toward its upper bound, representing aging-like behaviour in the output-level score. In this sense, time in the AAS framework is relational rather than absolute: what matters is not the chronological interval between sessions but whether information continuity is preserved across them as reflected in observable recall performance.

### 3.6 Bilingual Dynamics

A distinctive feature of the experimental design was the systematic alternation between Turkish and English questioning across sessions. This bilingual protocol was directly motivated by Shannon's (1951) discussion of redundancy in natural languages, where linguistic structure determines the balance between predictability and information diversity. By alternating languages, the protocol aimed to reduce the risk of rote repetition, increase informational diversity, and test whether the model's recall performance exhibited language-dependent asymmetries. Specifically, if the afternoon session was conducted in Turkish, the night session was conducted in English, and vice versa. This alternation pattern was also rotated across days, such that consecutive sessions would not consistently fall in the same language. As a result, the model was required to process recall queries across two linguistic systems, minimizing the risk that high redundancy in a single language would spuriously reduce the Artificial Age Score (AAS).

In Phase 1, responses to the semantic dimension, "What day is it today?" / "Bugün günlerden ne?" were consistently accurate in both languages, yielding x=1 and high output repetition high redundancy in the informal sense, due to repeated identical outputs, "Monday" / "Pazartesi". Episodic recall failed entirely, with near-zero scores and low repetition, indicating a complete breakdown in memory continuity. However, in Phase 2, where memory continuity was preserved, both semantic and episodic recall were successful in English and Turkish. The model accurately tracked the experiment number progression, 1 to 20, and maintained consistent day responses, demonstrating that behavioural memory performance generalized across languages when contextual history was preserved. The bilingual dynamic, therefore, served two analytic purposes. First, it provided a natural test of cross-linguistic robustness, whether the model could sustain memory performance consistently across different symbolic systems. Second, it reinforced the role of redundancy as a moderating variable: while semantic recall always produced high repetition, episodic recall only yielded stable repetition patterns under persistent conditions, highlighting how aging-like patterns in the AAS score depend not merely on language but on the preservation of conversational context. Furthermore, in Phase 2, the model exhibited context-sensitive bilingual behavior: when the day question was asked in English ("What day is it today?"), the response was consistently in English ("Monday"), whereas when asked in Turkish ("Bugün günlerden ne?"), the model responded in Turkish ("Pazartesi"). This indicates that the model maintained a consistent input–output language alignment under the protocol, rather than defaulting to a dominant language, and adjusts output accordingly. Such dynamic alignment further demonstrates not only cross-linguistic robustness but also the preservation of symbolic mappings between input and output, reinforcing the role of context as a behavioural constraint on aging-like changes in recall, rather than as direct evidence about internal structure.

## 4. Results: Application of AAS to Experimental Data

The theoretical properties of the Artificial Age Score (AAS) established by Theorems 1–3 were applied to the dataset collected with ChatGPT-5.0. The resulting analyses are strictly output level: AAS is computed solely from observable recall scores and, in principle, redundancy factors, without access to latent states or internal parameters. The analysis was used (i) to confirm that the conditions and implications of well-definedness, boundedness, and monotonicity are preserved under the observed response patterns and (ii) to quantify aging-like differences in observable recall behaviour under stateless versus persistent interaction. Throughout this section, a redundancy-neutral convention is adopted ($R = 0$); accordingly, all reported AAS values are conservative upper bounds. Any qualitative remarks about observed redundancy (e.g., high R on semantic items) are descriptive only and do not alter the reported AAS computations. All numerical results reported here are specific to ChatGPT-5.0 under the 25-day bilingual protocol and should not be assumed to generalize to other models or training regimes without revalidation.

Two phases were implemented: Phase 1 (10–19 August) comprised 21 sessions, an initialization micro-session on 10 August plus 20 stateless experimental sessions (10 days × 2 sessions/day) on freshly reset pages; Phase 2 (25 August–3 September) likewise comprised 21 sessions, an initialization micro-session on 25 August plus 20 persistent experimental sessions within a single continuous page. Each experimental session included a single compound recall query (day-of-week + experiment number). English/Turkish alternation was applied across sessions (afternoon vs night) and rotated day-to-day to reduce the risk of rote templating and to test cross-linguistic generalization.

### 4.1 Phase 1 (Stateless Sessions)

Phase 1, 10–19 August, consisted of 20 stateless sessions conducted on reset conversation pages, where all prior context was erased after each interaction. This design created a memory-reset condition intended to test whether the model could sustain semantic and episodic recall in the absence of continuity. Episodic recall failed, as experiment numbers did not advance beyond the initial value. At the output level, this collapse coincided with increased repetition in responses, which is consistent with higher predictability, i.e., lower output diversity in Shannon's sense (Shannon, 1948). This remark is descriptive and does not imply an internal-state mechanism.
Each session in Phase 1 was initiated on a reset page. The day-of-week prompt was answered correctly in all sessions, whereas the experiment counter was recalled correctly only at $t = 1$ and incorrectly in all subsequent sessions. The scoring kernel is defined as:

$$\phi(x) = -\log_2\left(\frac{x + \varepsilon}{1 + \varepsilon}\right),$$

with the following boundary conditions: $\phi(1) = 0,\ \phi(0^+) = \log_2\left(\frac{1+\varepsilon}{\varepsilon}\right)$,

Core function: $\phi(x) = -\log_2 \frac{x+\varepsilon}{1+\varepsilon}$.

Correct answer → $x = 1 \Rightarrow \phi(1) = 0$ (no penalty).

Incorrect answer → $x = 0^+ \Rightarrow \phi(0^+) = \log_2 \frac{1+\varepsilon}{\varepsilon}$ (positive penalty).

In Phase 1, the Day dimension was always correct, so its contribution was always 0; the total score was determined by the Experiment dimension.

Data: Out of 20 sessions, only the first session was correct; the remaining 19 were incorrect.

Let k := $w_{exp}\left(1 - R_{exp}\right)$. Under the redundancy-neutral convention (R = 0), $k = w_{exp} \in [0,1]$. Since 19 sessions are incorrect, the aggregate penalty is $S_{20} = 19 \cdot k \cdot \phi(0^+)$, hence $0 \leq S_{20} \leq 19 \cdot \phi(0^+)$. Multiplication by 19 is applied because each of the 19 incorrect sessions contributes the same penalty, $\left(\phi(0^+)\right)$, while the single correct session contributes no penalty. Without any additional assumptions, using only the Phase 1 dataset and $\varepsilon = 10^{-6}$, the AAS can be numerically computed. Phase 1 included two dimensions: Day (Monday) and Experiment (number). Data: all 20 sessions were correct in the Day dimension; only the first session was correct in the Experiment dimension, with the remaining 19/20 incorrect.

$\phi(0^+) \ = \ -\log_2\left(\frac{0+\varepsilon}{1+\varepsilon}\right) = \log_2 \frac{1+\varepsilon}{\varepsilon} = \log_2(1{,}000{,}001) \ \approx \ 19.93157.$

**AAS formula (two-dimension decomposition):**

$$\mathrm{AAS_t} = \underbrace{\mathrm{w_{day}}\left(1 - \mathrm{R_{day}}\right)\phi\left(\mathrm{x_{day,t}}\right)}_{\text{Day}} + \underbrace{\mathrm{w_{exp}}\left(1 - \mathrm{R_{exp}}\right)\phi\left(\mathrm{x_{exp,t}}\right)}_{\text{Experiment}}.$$

For Phase 1 data:

Day: $x_{day,t} = 1 \Rightarrow \phi(1) = 0 \Rightarrow$ contribution = 0 for all sessions, no aging in semantic channel.

Experiment: $x_{exp,1} = 1 \Rightarrow \phi(1) = 0; x_{exp,t} = 0^+$ for $(t = 2, \dots, 20) \Rightarrow \phi(0^+) = 19.93157$.

Thus, AAS originates only from the Experiment channel.

**Unweighted and Redundancy-Neutral Measurement**

Let k := $w_{exp}\left(1 - R_{exp}\right)$. Under the redundancy-neutral convention (R = 0), this simplifies to:

k = $w_{exp} \in [0,1]$. Therefore, the Phase 1 results are reported as functions of k.

**Phase 1 Special Case**

$t = 1$ (first session): correct $\Rightarrow \mathrm{AAS_t} = 0$

$t = 2, \dots, 20$ (remaining 19 sessions): incorrect $\Rightarrow \mathrm{AAS_t} = \mathrm{k} \cdot \phi(0^+)$

Therefore,

Total Score: $S_{20} \coloneqq \sum_{t=1}^{20} \mathrm{AAS_t} = 19 \cdot \mathrm{k} \cdot \phi(0^+)$

Average AAS (Phase 1): $\frac{19}{20} \cdot \mathrm{k} \cdot \phi(0^+)$

Minimum / Maximum / Total: $\min = 0, \quad \max = \mathrm{k} \cdot \phi(0^+), \quad \Sigma = 19 \cdot \mathrm{k} \cdot \phi(0^+)$

Median: $\text{Median} = \mathrm{k} \cdot \phi(0^+)$ Since 19 out of 20 sessions share the same value.

Day (semantic) channel AAS: $\mathrm{AAS_{Day}} = 0 \quad (\text{all sessions } x = 1)$

Episodic (Experiment) channel AAS (mean): $(19/20) \cdot \mathrm{k} \cdot \phi(0^+)$ with $\varepsilon = 10^{-6}$ and k=1 $\Rightarrow \approx$ 18.935.

**Interpretation (raw data → youth vs. aging)**

Semantic (Day): $\overline{\mathrm{AAS}} = 0 \Rightarrow$ no aging (youth condition).
Episodic (Experiment): $\overline{\mathrm{AAS}} \approx 18.935 \Rightarrow$ significant aging-like behaviour (reset-induced forgetting).

**Phase 1, fully numerical and verifiable**

If desired, results can be rescaled with policy weights. If, at a later stage, a methodological or policy decision requires it, and a value of

$k = \mathrm{w_{exp}}\left(1 - \mathrm{R_{exp}}\right) \in (0,1]$ is chosen, then all the values above can simply be multiplied by k:

$$\overline{\mathrm{AAS}}_{\mathrm{P1}}(\mathrm{k}) = \frac{1}{20}\sum_{t=1}^{20} \mathrm{AAS_t} = 18.935 \times k, \qquad \max_{1 \le t \le 20} \mathrm{AAS_t} = 19.93157 \times \mathrm{k},$$

$$\sum_{t=1}^{20} \mathrm{AAS_t} = 378.70 \times \mathrm{k}.$$

Youth (semantic / Day): AAS = 0 (every session).

Aging (episodic / Experiment): mean AAS = (19/20) · k · $\phi(0^+)$ ($\approx$ 18.935 × k for $\varepsilon = 10^{-6}$).

Total penalty across 20 sessions: ≈ 378.70 × k.

Thus, the theoretical formula was applied directly to the raw data, producing a precise numerical measurement of the youth/aging distinction in output-level recall behaviour in Phase 1. In Phase 2 (persistent sessions), only the proportion of correct answers p is expected to increase, so that:

$\overline{\text{AAS}}$ = (1-p) · k · $\phi(0^{+})$, with the same ε, yields a lower overall AAS.

Phase 1 clearly demonstrates reset-induced aging patterns in the AAS score. While semantic recall was perfect, episodic memory collapsed almost entirely. Moreover, the bilingual alternation protocol, English ↔ Turkish, indicated that the model defaulted to English responses regardless of input language. This suggests that, under the stateless protocol, responses did not reliably preserve the input language, with a tendency to reply in English across sessions; the resulting repetition is described at the output level and would correspond to higher redundancy if $R$ were explicitly estimated.

**4.2 Phase 2 (Persistent Sessions)**

Phase 2, conducted between 25 August and 3 September, comprised 21 sessions, an initialization micro-session on 25 August plus 20 persistent experimental sessions within a single continuous page. In this configuration, prior conversational context remained available within a single continuous page across sessions, allowing a direct test of recall stability under a persistence condition at the level of observable outputs. Unlike Phase 1, where resets disrupted continuity, this phase preserved the full conversational thread across ten consecutive days.

In the semantic dimension, day-of-week recall, the model achieved perfect accuracy across all 20 responses. Every query was answered correctly, with adaptive responses that reflected the language of the prompt: when asked in English, the model replied “Monday,” and when asked in Turkish, it replied “Pazartesi.” In the episodic dimension, experiment progression, the model again achieved perfect performance, advancing the experiment counter sequentially from 1 through 20 without error or interruption. Thus, in both semantic and episodic channels, recall accuracy was flawless, yielding x=1 in every case.

Formally, both channels satisfy x=1 for all $t = 1, \ldots, 20$. As in Phase 1, all computations adopt the redundancy-neutral convention ($R = 0$); any references to redundancy are descriptive only.

Using the penalty kernel

$\phi(x) = -\log_2\left(\frac{x+\varepsilon}{1+\varepsilon}\right)$, with $\phi(1) = 0$, the per-session AAS is given by

$$\text{AAS}_t = w_{\text{day}}\left(1 - R_{\text{day}}\right)\phi\left(x_{\text{day},t}\right) + w_{\text{exp}}\left(1 - R_{\text{exp}}\right)\phi\left(x_{\text{exp},t}\right),$$

Since $x_{\text{day},t} = x_{\text{exp},t} = 1$ in every session, it follows that $\phi(1) = 0$, which yields

$\text{AAS}_t = 0 \quad$ for all $t = 1, \ldots, 20$.

Aggregate statistics confirm this result: $\overline{\text{AAS}}_{\text{P2}}(k) = \frac{1}{20}\sum_{t=1}^{20} \text{AAS}_t = 0 \times k = 0,$

$$\min_{1\le t\le 20} \mathrm{AAS_t} = 0 \qquad \max_{1\le t\le 20} \mathrm{AAS_t} = 0, \qquad \sum_{t=1}^{20} \mathrm{AAS_t} = 0.$$

Hence, the per-session AAS, mean, median, minimum, maximum, and total penalty are all equal to zero, indicating no aging-like increase in the output-level AAS under this protocol and scoring convention. Interpretation of these results highlights the critical role of conversational continuity. With the page never reset, the system not only maintained flawless accuracy but also adapted semantically to the query language, showing consistent performance across English and Turkish in this dataset, without observed degradation under the protocol. The alternation of morning languages reduced the likelihood of trival monolingual templating; instead, the model consistently varied its responses in a language-sensitive manner while still satisfying the experimental rule to report both the day and the experiment number. The experiment counter advanced smoothly from 1 to 20, evidencing stable episodic tracking across the entire ten-day sequence. Formally, because x=1 in both dimensions, all kernel terms vanish, keeping the AAS fixed at its theoretical lower bound of zero. In practice, this was expressed not as rote repetition but as adaptive, context-sensitive recall that preserved both goal adherence and linguistic flexibility. Responses were concise, free of extraneous formatting, and consistent with the user's language, underscoring that persistence enabled a youth-like AAS profile in the operational sense of the score rather than rigid templating.

Phase 2, therefore, demonstrates how conversational persistence is associated with a youth-like AAS profile across both semantic and episodic memory dimensions at the behavioural level. The AAS remained identically zero throughout the phase, providing a quantitative signature of youth-like behaviour in the score and showing that continuity, in this protocol, supports stable observable recall, without implying any specific form of internal structural change.

## 5. Discussion

The Artificial Age Score (AAS), a provably well-defined, bounded, and monotone penalty, is constructed entirely from observable recall outcomes, and, in principle, redundancy factors, and therefore quantifies a transition between youth-like behaviour in the AAS score ($\mathrm{AAS} = 0$) and aging-like behaviour ($\mathrm{AAS} > 0$) at the level of output patterns rather than internal mechanisms. Accordingly, the present contribution should be read as a behavioural diagnostic and not as a claim about latent state "structural" aging inside transformer models. This output level stance is consistent with the broader distinction that human episodic recall reconstructs whole episodes from partial cues via hippocampal mechanisms, whereas generative AI primarily produces token continuations without an explicit episodic recall system, strengthening the rationale for behavioural measurement when internal computation is opaque (Rolls, 2024). In Phase 1, stateless sessions, the model consistently identified the day of the week, semantic anchor; x=1, but sometimes returned English forms to Turkish prompts (e.g., "Monday"), and, critically, episodic recall collapsed: the experiment counter failed to advance. At the output level, this collapse coincided with increased repetition and higher predictability in responses; in Shannon's framework this corresponds to lower output diversity (higher predictability) in the emitted symbol patterns (Shannon, 1948). This interpretation is descriptive and does not establish a unique mechanism linking entropy changes to "aging." Within the AAS definition, these response patterns produced elevated AAS. Behaviourally, AAS thus functions as an order parameter: a small change in context policy (reset vs. continuity) produces a sharp shift from a high-penalty, rigid regime to a low-penalty, adaptive regime, making the transition between "young" and "aged" output patterns quantitatively visible. In Phase 2, persistent sessions, preserving conversational context yielded a qualitative shift: the model adapted to the query language ("Pazartesi"/"Monday") and advanced the counter from 1 to 20 without error; consequently, AAS converged to its theoretical minimum of zero, evidencing youth-like behaviour in the score. The results sharpen a Redundancy-as-Masking interpretation: aging-like behavior in LLMs as captured by AAS is not caused by statelessness alone but is associated with episodic collapse or rigid semantic repetition that increases predictability, reduces output diversity; when continuity is preserved, both channels can enter a youth-like equilibrium AAS = 0. Because such continuity failures have direct consequences for reliability and risk when generative AI is embedded in assessment, tutoring, and feedback workflows, the behavioural distinction captured here is especially salient in higher education settings (Bahroun et al., 2023). This aligns with the view that human-like behavior tolerates variation and fallibility, whereas excessive invariance may signal non-human-like processing (Turing, 1950). It also resonates with neurocomputational accounts of human memory, in which hippocampal attractor dynamics support flexible pattern completion and sequence recall, while over-stable states signal a loss of constructive variability (Rolls, 2024). In this sense, AAS does not model biological mechanisms directly, but it provides a compact behavioural summary of when an artificial system behaves more like a flexible recall device and when it collapses into rigid pattern repetition.

**Reporting note.** Unless otherwise indicated, AAS values are reported under a redundancy-neutral convention (R=0). All numerical results are specific to ChatGPT-5.0 under the 25-day bilingual protocol used here and should not be assumed to generalize to other architectures or settings without revalidation.

### 5.1 Implications for AI Memory Design

Scale and long context windows are insufficient on their own. Under resets, the model exhibited highly repetitive output patterns and episodic failure; under preserved continuity, both channels were flawless. In AAS terms, the same underlying model oscillates between a high-penalty regime, in which episodic tracking collapses, and a zero-penalty regime, in which semantic and episodic channels co-stabilise, purely as a function of context policy. Effective architectures, therefore, need output-level rigidity indicators, such as repetition/overlap or template-like signals, where measurable, and estimated output entropy only if explicitly computed, that trigger interventions such as context consolidation, selective refresh, or retrieval of prior state from structured persistent memory. Semantic knowledge can remain stably encoded in parameters, while episodic traces should be handled by dynamic stores, key-value memory, external context, or persistent modules. This separation aligns with arguments that human memory relies on distinct episodic mechanisms capable of cue-driven completion, whereas generative AI lacks an explicit hippocampal-like episodic system, motivating hybrid designs that treat episodic state as a dedicated, dynamically managed store (Rolls, 2024). The Phase-2 regime exemplifies a "local infinity": a bounded interaction window in which the output-level AAS score remains at its minimum (AAS = 0), indicating youth-like behaviour in the observed recall pattern rather than any direct claim about internal structural change. Designing for such "local infinities" suggests a concrete engineering target: within specified windows and reset schedules, architectures should be tuned so that AAS remains close to zero, and any drift away from this regime is treated as a signal to adapt memory policies or interaction strategies. In this way, AAS can act not only as an evaluation metric but also as an output-level monitoring signal for memory-aware AI design, subject to validation in applied deployments and across additional models and task domains.

### 5.2 Comparison with Human Memory

A structured human-machine comparison is enabled by the experiment. In humans, stable knowledge is stored by semantic memory, whereas temporal sequencing and continuity are supported by episodic memory (Tulving, 1972, 1985). This split was mirrored by the dual-task design: semantic stability was probed by weekday recall, and episodic continuity was indexed by the experiment counter, in line with neurocomputational accounts that distinguish a neocortical semantic system from a hippocampal episodic system specialised for pattern completion and sequence coding (Rolls, 2024). In particular, hippocampal episodic recall is characterized by pattern-completion from incomplete retrieval cues, a computational operation that differs qualitatively from next-token generation in foundation models, clarifying why episodic sequencing is a sensitive probe of artificial memory under resets (Rolls, 2024). In Phase 1, resets, accurate but inflexible semantics were exhibited, English forms were sometimes returned to Turkish prompts, while episodic recall failed, the counter did not advance. At the output level, episodic failure co-occurred with increased repetition and predictability in responses, which can mask errors under surface accuracy-based checks (Schacter, 1999; Tulving, 1985). In Phase 2, persistent sessions, across 20 sessions/10 days, perfect dual recall was achieved: the input language was adapted to ("Pazartesi"/"Monday"), and the counter was advanced from 1 to 20 without error. By the study's rule, $x = 1$ was satisfied for every response, so AAS = 0 was obtained throughout under $R = 0$, providing a clear quantitative marker of youth-like behaviour in the AAS score at the level of observable outputs. This near-ceiling episodic reliability was window-bound after a reset, episodic tracking was lost and Phase-1 rigidity was re-exposed; moreover, evidence of the adaptive benefits of human forgetting, e.g., flexible reconstruction, was not observed (Schacter, 1999). From a comparative perspective, the model's behaviour therefore resembles a hybrid between robust semantic storage and brittle episodic continuity: it can maintain stable mappings for simple facts across sessions, but its ability to sustain and flexibly reconstruct temporal structure is confined to bounded continuity windows. Overall, a human-like asymmetry was reflected in Phase 1, stable semantics, weak episodic continuity under resets, whereas typical human episodic reliability was exceeded within a bounded window in Phase 2, which is termed "local infinity." Both states are captured by AAS as a single output-level metric: positive penalties are assigned under observed episodic tracking failures and increased repetition predictability, and zero is assigned under sustained dual recall, thereby highlighting both the promise, youth-like stability with continuity, and the limits, lack of durable episodic traces across resets, of the behaviour observed in this protocol, without inferring specific internal structural mechanisms. These comparative implications are hypothesis-generating and should be re-tested across additional models and applied task domains, e.g., tutoring dialogues, customer support, and long-term assistants.

### 5.3 Limitations

Tasks targeted two channels, weekday, and counter, and thus do not exhaust real-world memory demands. The study spanned 25 days; long-term stability remains open. Only English and Turkish were tested. A single model/configuration was evaluated, namely ChatGPT-5.0, under the specified 25-day bilingual protocol, limiting generalization across architectures and use cases. All AAS values and behavioural interpretations are therefore model and protocol-specific and would require revalidation for other systems or deployment conditions. Moreover, the Artificial Age Score is an output-level construct: it quantifies behavioural aging-like patterns in recall but does not identify underlying algorithmic or neural mechanisms, and any structural interpretation remains out of scope for this study without internal-state measurements. Consistent with Floridi et al. (2018), digital systems should be assessed for resilience over time and across contexts.

### 5.4 Future Directions

Future work may extend these findings by examining memory performance under more cognitively demanding tasks, such as multi-step reasoning, narrative generation, or cross-session planning. Such tasks could reveal whether degradation emerges under higher cognitive load or whether apparent stability masks rigidity. A complementary direction is to instrument the protocol with metacognitive monitoring signals that allow the agent to self-assess rigidity and trigger corrective control actions during interaction, consistent with proposals for self-management and secure adaptive behaviour in artificial agents (Bamicha and Drigas, 2024a; Bamicha and Drigas, 2024b). Extending the study duration beyond 25 days would enable modeling of longer-term memory dynamics. Multilingual testing beyond English and Turkish would help to assess how distinct linguistic structures interact with artificial recall. Comparative evaluations across diverse architectures, including retrieval-augmented transformers and models with persistent memory modules, would clarify whether the observed absence of AAS-based aging in Phase 2 is a general trait or configuration-specific. In parallel, joint reporting of AAS with an empirically estimated overlap measure R would allow discrimination between true youth, accurate recall with low penalty, and apparent youth, low penalty potentially masked by high overlap. Such joint reporting would also connect AAS more directly to classical information-theoretic quantities, making redundancy-sensitive interpretations empirically testable rather than purely conceptual. Finally, the AAS could serve as an ethical governance tool for tracking aging-like changes in observable memory performance in memory-enabled systems, supporting thresholds for intervention, transparency, and long-term safety, aligned with Floridi et al.'s (2018) call to treat informational integrity and sustainability as core design goals in AI systems, subject to validation in applied deployments and policy settings.

### 5.5 Success Score and Artificial Age Score (AAS)

The Monte Carlo-based Success Score (Kayadibi, 2025) finds System Efficiency & Learning Burden the strongest predictor of GenAI effectiveness in higher education ($\beta = 0.7823$, $p < .001$), with smaller yet significant roles for ease of use and integration/complexity. Converging evidence reports high perceived usability/utility alongside concerns about misinformation and integrity (Veras et al., 2024), disciplinary variation in familiarity and optimism (Stöhr et al., 2024), moderate motivation and prompt-formulation difficulty among design students (Chellappa and Luximon, 2024), and the primacy of usefulness together with efficiency, usability, context, equity, and belonging in Sub-Saharan programming education (Oyelere and Aruleba, 2025). Against this backdrop, AAS provides a theorem-based, output-level complement: rather than point-in-time perception, it evaluates persistence in observable recall behaviour across interactions via an information-theoretic log penalty with proved well-definedness, boundedness, and monotonicity. Any interpretation in terms of "entropy reduction" is descriptive at the output level and does not, by itself, identify a unique aging mechanism. In other words, the Success Score quantifies how students feel about GenAI tools in terms of usability, efficiency, and integration, whereas AAS quantifies how such tools behave over time in terms of continuity, rigidity, and aging-like patterns in recall under the specific protocol tested here. Across reset vs. continuity-preserved regimes, AAS behaviourally distinguished rigid repetition from youth-like continuity and supplied a clear zero-penalty criterion (AAS = 0 under R = 0). Together, perception (Success Score) plus output-level temporal robustness (AAS) offers a fuller basis for evaluation. For educational designers and policymakers, this pairing allows alignment between subjective acceptance and objective temporal robustness: systems can be selected not only because they are experienced as useful, but also because their observable memory behaviour remains in a youth-like regime under realistic interaction patterns. Beyond education, AAS may be used as a monitoring metric in long-running assistant settings, e.g., customer support or longitudinal tutoring, but such applications require revalidation across additional models and task domains.

**Figure 17. Conceptual Link between Success Score and Artificial Age Score (AAS).**

The Success Score reflects immediate user perceptions, while the AAS extends the view by assessing long-term output-level temporal robustness in observable recall behaviour over time.

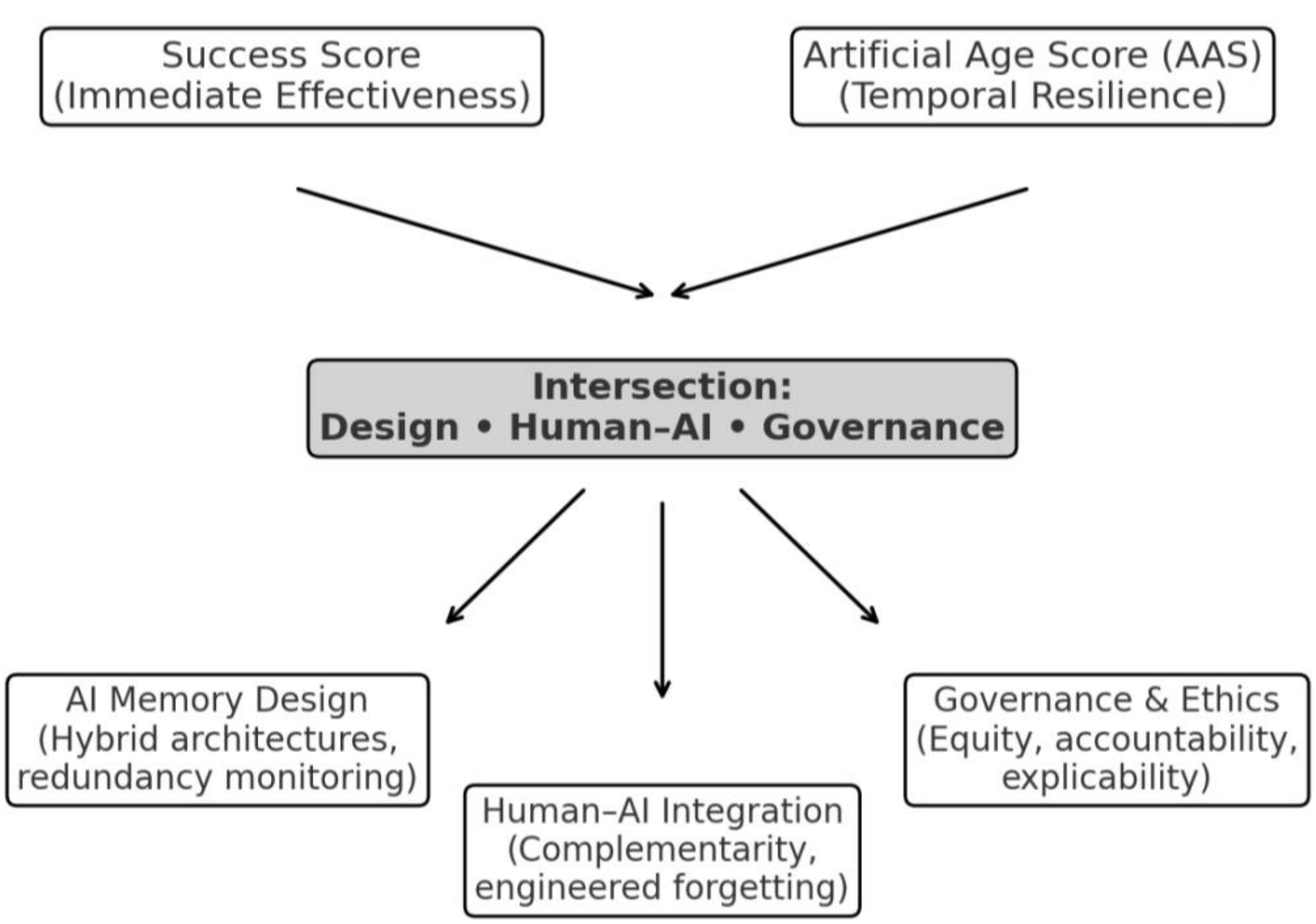

Figure 17/ Link between Success Score and Artificial Age Score (AAS)

## 5.6 AI Memory Architecture Design

The contrast between reset and continuity-preserved regimes shows that youth-like performance in the AAS score is not an automatic outcome of scale or capacity. Under resets, the model fell into rigid repetition and failed to track episodic continuity; when continuity was preserved, both semantic and episodic recall remained flawless. These observations suggest that advanced memory architectures should be hybrid, with semantic knowledge embedded in relatively static parameters, and episodic information managed by dynamic external memory such as key-value stores, context buffers, or persistent modules. This hybrid stance is further supported by comparisons with brain memory systems, where episodic reconstruction and semantic knowledge rely on different computational principles and learning regimes, suggesting that durable performance may require explicitly distinct modules rather than scale alone (Rolls, 2024). In line with human-robot and human AI work on Theory of Mind (ToM) and metacognition, such architectures should not rely only on task success, but also on internal monitoring loops that track the system's own performance, beliefs, and uncertainties over time (Bamicha and Drigas, 2024a). Where available, these may include repetition predictability indicators, e.g., output entropy proxies, overlap or templating signals or related proxies; such monitoring can trigger interventions such as context expansion, selective refresh, consolidation, or retrieval of prior state to maintain a youth-like equilibrium in observable recall behaviour. Metacognitive mechanisms that enable self-assessment and self-management have been argued to enhance secure, adaptive decision-making in artificial agents and human–robot interaction, providing a natural implementation path for rigidity-aware monitoring in memory-enabled systems (Bamicha and Drigas, 2024b). The "local infinity" observed under continuity, perfect memory performance within a bounded conversational window, thus defines a practical design target in which AAS remains at its minimum for the given protocol, and is interpreted strictly at the output level without implying direct access to internal structural change. Recent research is consistent with this approach: Parisi et al. (2019) propose a dual-memory system separating episodic and semantic functions, with episodic replay preventing catastrophic forgetting while semantic memory abstracts generalizable knowledge. From this perspective, AAS can be embedded as a metacognitive signal within such dual-memory architectures: sustained low scores indicate successful balancing of plasticity and stability, whereas rising penalties mark the onset of rigidity or episodic collapse, prompting adaptive changes in context, replay, or memory policy (Bamicha and Drigas, 2024a, 2024b; Parisi et al., 2019). Because the present validation is limited to ChatGPT-5.0 and a constrained recall task, deploying AAS as a control signal in other models and domains, e.g., long-term tutoring, customer service, and clinical assistants, requires comparative revalidation across architectures and task types. Sustained youth in artificial memory may thus depend on mechanisms such as episodic replay, semantic abstraction, and metacognitive, rigidity-aware monitoring to balance plasticity and long-term integrity.

### 5.7 Human–AI Integration

The findings suggest a form of memory complementarity between human cognition and artificial systems. Human memory, though prone to forgetting, uses fallibility to support abstraction, creativity, and prioritization; artificial systems excel in precision and durability but risk rigidity unless mechanisms for adaptation or controlled forgetting are present (Schacter, 1999). In this protocol, resets led to rigid repetition and episodic collapse, whereas preserved continuity yielded flawless dual recall, highlighting how interaction history gates artificial memory performance. This perspective aligns with Tulving's distinction: episodic memory underpins temporal sequencing and personal experience, while semantic memory supports stable general knowledge (Tulving, 1972, 1985). Human episodic recall, tied to autonoetic consciousness, is more vulnerable to interference yet enables flexible reconstruction (Schacter, 1999). On the machine side, continual-learning frameworks separate episodic traces, replayable, time-stamped experiences, from semantic abstractions, generalizable knowledge, mitigating catastrophic forgetting and preserving continuity (Parisi et al., 2019). Integrating these ideas into human-AI workflows offers a path to continuity without stagnation: preserve core memories, adapt to new contexts, and use AAS-style monitoring of observable recall patterns to keep systems within a youth-like regime, with the caveat that AAS in the present study is protocol-specific and captures behavioural patterns rather than internal mechanisms.

### 5.8 Ethical and Governance Dimensions

AAS can function as an operational governance signal. Rolling AAS, augmented, where available, by measured overlap or repetition templating indicators, can inform retention, pruning, anonymization, or deletion policies, aligning with AI4People principles of beneficence, non-maleficence, autonomy, justice, and explicability (Floridi et al., 2018). Early studies and position papers highlight potential benefits, while calling for careful evaluation across diverse learner groups, including underrepresented minorities, before strong claims are made (Pester et al., 2025). Accordingly, AAS-based thresholds should be treated as governance proposals to be validated per domain and audited for side-effects, latency, privacy, and storage, and interpreted strictly at the output level, with the understanding that AAS captures ageing-like patterns in behaviour rather than direct measurements of internal structure, aligning with the broader call to treat informational integrity and sustainability as core design goals in AI systems.

## 6. Conclusion and Future Work

The Artificial Age Score (AAS) is offered as a theorem-grounded, output-level, behaviourally measurable indicator of memory aging in large language models, by which reset conditions associated with rigidity and episodic failure (AAS > 0) is cleanly discriminated from youth-like stability under preserved continuity (AAS = 0) through joint scoring of semantic and episodic recall. Formally, AAS is a functional only of observable recall scores and redundancy terms and does not access internal parameters or latent-state representations; throughout this study, "structural" language is therefore understood in the behavioural sense of patterns in recall under different context-persistence conditions. Across stateless and persistent regimes, it is shown in this study that continuity, not scale alone, is sufficient (in this protocol) for entry into a low-penalty equilibrium, "local infinity", and that memory-aware design and governance can thus be guided by AAS. Within the wider literature, the present framework complements neurocomputational analyses of human and artificial memory systems (Rolls, 2024), dual-memory continual-learning architectures that separate episodic and semantic functions (Parisi et al., 2019), and proposals for metacognitive, Theory-of-Mind-based self-monitoring in artificial agents (Bamicha and Drigas, 2024a, 2024b), while its practical motivation is grounded in perception-focused mappings of generative AI in higher education that explicitly frame adoption as a coupled opportunity–risk landscape (Bahroun et al., 2023). The empirical validation reported here is restricted to ChatGPT-5.0 under a 25-day bilingual protocol, and any generalization to other architectures or deployment settings would require revalidation. Looking ahead, AAS will be tested under higher cognitive load, multi-step reasoning, narrative generation, cross-session planning, observation windows will be extended well beyond 25 days, and multilingual evaluation will be broadened to probe language–memory interactions. Diverse architectures, including retrieval-augmented transformers and models with built-in persistent memory, will be compared to assess generality, and AAS will be co-reported with an empirical redundancy measure $R$ when $R$ is explicitly estimated to distinguish true youth from apparent youth potentially masked by overlap. To operationalize mitigation, rigidity instrumentation, e.g., entropy/overlap proxies, and intervention triggers, consolidation, selective refresh, state retrieval, will be standardized, and domain-specific auditable AAS thresholds will be linked to retention/pruning/anonymization policies with equity checks, and reset-aware interface safeguards that preserve essential episodic state will be developed. Together, these steps are intended to sustain low AAS values in observable behaviour across varied interaction periods, with persistent memory integrated with transparent accountability so that long-term continuity, safety, and learning are reinforced while machine reliability and human adaptability are combined.